\documentclass[]{article}

\usepackage[utf8]{inputenc} 
\usepackage[T1]{fontenc}    
\usepackage{lmodern}
\usepackage[hidelinks]{hyperref}       
\usepackage{url}            
\usepackage{booktabs}       
\usepackage{amsfonts}       
\usepackage{nicefrac}       
\usepackage{microtype}      
\usepackage[numbers]{natbib}

\usepackage{mathtools} 
\usepackage{enumitem}
\setlist[itemize]{topsep=0pt, itemsep=-1ex,partopsep=1ex, parsep=1ex}
\usepackage{amsmath}
\usepackage{graphicx}
\usepackage{subfigure}
\usepackage{caption}
\usepackage{color}
\usepackage[dvipsnames]{xcolor}
\usepackage{authblk}

\newcommand{\E}{\mathbb{E} \,} 
\newcommand{\R}{\mathbb{R}} 
\newcommand{\grad}{\nabla} 

\newcommand{\commentout}[1]{}

\DeclareMathOperator{\CoV}{Cov} 
\DeclareMathOperator{\Var}{Var} 

\DeclareMathOperator{\diag}{diag}
\DeclareMathOperator{\tr}{tr} 
\DeclareMathOperator*{\argmin}{argmin} 
\DeclareMathOperator*{\argmax}{argmax} 

\newcommand{\presec}{\vspace*{-0pt}}
\newcommand{\postsec}{\vspace*{-0pt}} 
\newcommand{\pressec}{\vspace*{-0pt}}
\newcommand{\postssec}{\vspace*{-0pt}}
\newcommand{\prepar}{\vspace*{-0pt}}

\newcommand{\preeq}{\vspace*{-0pt}}
\newcommand{\posteq}{\vspace*{-0pt}}
\newcommand{\preitem}{\vspace*{-0pt}}
\newcommand{\postitem}{\vspace*{-0pt}}

\begin{document}

%

%
\title{Approximate Collapsed Gibbs Clustering with Expectation Propagation}

\author{Christopher Aicher\thanks{Department of Statistics, University of Washington, \texttt{aicherc@uw.edu}} \hspace{0.3em}and Emily B. Fox\thanks{Department of Computer Science and Statistics, University of Washington, \texttt{ebfox@uw.edu}}}

%

\date{}
\maketitle

\begin{abstract}
    We develop a framework for approximating collapsed Gibbs sampling
    in generative latent variable cluster models. 
    Collapsed Gibbs is a popular MCMC method,
    which integrates out variables in the posterior
    to improve mixing.
    Unfortunately for many complex models,
    integrating out these variables is either
    analytically or computationally intractable.
    We efficiently approximate the necessary collapsed Gibbs integrals by
    borrowing ideas from expectation propagation.
    We present two case studies
    where exact collapsed Gibbs sampling is intractable:
    mixtures of Student-$t$'s and time series clustering.
    Our experiments on real and synthetic data show that our approximate
    sampler enables a runtime-accuracy tradeoff in sampling these types of models,
    providing results with competitive accuracy much more rapidly than
    the naive Gibbs samplers one would otherwise rely on in these scenarios.
\end{abstract}

\presec
\section{Introduction}
\postsec

A common task in unsupervised learning is to cluster observed data 
into groups that are similar.
One principled approach is to infer latent cluster assignments in a hierarchical probabilistic model.
Hierarchical latent variable models have the benefit of allowing for both
(i) more flexible and complex models to be built from simpler distributions and
(ii) statistical strength to be shared within clusters for inference.
Examples of latent variable models for clustering include
mixture models~\cite{escobar1995bayesian,dunson2000bayesian},
topic models~\cite{blei2003latent, blei2007correlated},
and network block models~\cite{snijders1997estimation, airoldi2008mixed}.
However, a key obstacle in fitting these latent variable models 
is searching over the combinatorial number of different 
clustering assignments.

For simple conjugate models,
a variety of methods have been proposed for Bayesian inference of
the latent cluster assignments, including
variational inference~\cite{blei2017variational} and
Markov chain Monte Carlo (MCMC)~\cite{bishop2006pattern}.
In this paper, we focus on MCMC and present an approximation algorithm in a similar spirit to other recent approximate MCMC techniques (cf.,~\cite{johndrow2015approximations, murray2004bayesian}).  
Although variational methods have seen great advances recently, proving quite powerful and scalable~\cite{hoffman2013stochastic}, there are still known drawbacks such as underestimation of uncertainty,
a key quantity in a full Bayesian analysis. 

In terms of MCMC methods, the simplest is Gibbs sampling,
which iteratively draws individual cluster assignments and model parameters
from the posterior conditioned on all other variables. 
While such \emph{naive} Gibbs sampling is theoretically guaranteed to converge,
in practice, it 
is known to mix slowly in high dimensions~\cite{van2008partially}.
A popular modification is \emph{collapsed} Gibbs sampling,
which iteratively draws from marginals of the posterior
by integrating out variables.
Integrating out variables reduces the dimension of the posterior and
often eliminates local modes arising from tightly coupled variables~\cite{liu2008monte}. 
Unfortunately, for complex models, sampling from the marginal posterior can be 
analytically intractable or computationally prohibitive. 

For example, in the time series clustering model of 
\citet{ren2015achieving}, collapsed Gibbs sampling requires running
a computationally intensive Kalman smoother per iteration
that scales cubically in the number of series per cluster.
Another common example is mixture modeling with non-conjugate emissions.
One example is the Student-$t$, which is popular in robust modeling 
due to its ability to capture heavy-tails.
In such cases, the emission parameters cannot be directly integrated out due to non-conjugacy.
In these two cases,
which we use as illustrative of the challenges faced in many models appropriate for real-data analyses,
collapsed Gibbs sampling is either infeasible or impractical.

A recently popular alternative MCMC technique to Gibbs sampling is the class of Hamiltonian Monte Carlo (HMC)-like algorithms~\cite{neal2011mcmc} and their scalable variants (cf.,~\cite{ma2015complete}).  These algorithms utilize (stochastic) gradient information about the target posterior and simulate continuous dynamics to efficiently explore the distribution. However, these methods only apply to fixed-sized continuous parameter spaces.  In our setting, the discrete latent cluster indicator variables must be marginalized out. The resulting non-conjugate marginalized log-likelihood terms can be handled using auto-differentiation. However, these methods require handling ``label switching", do not apply to nonparametric mixtures, and are slow for large clusters with complex likelihoods.  As such, this class of MCMC techniques does not maintain the spirit of collapsed Gibbs.
One such approximately collapsed method is `griddy Gibbs'; however it is limited to univariate variables~\cite{ritter1992facilitating}.

We instead stay within the collapsed Gibbs framework and aim to address how to handle the challenging required integrals in many scenarios.
We draw inspiration from expectation propagation (EP)~\cite{minka2001expectation,seeger2005expectation} and approximate the intractable integrals in cases where moments can be matched. 
Traditionally, EP is a method of approximating a target distribution with a distribution from a fixed simpler family of distributions, usually an exponential family.
In our case, instead of using EP to directly approximate the posterior of cluster assignments~\cite{fan2014non},
we use EP to approximate the conditional posterior of the nuisance parameters we wish to collapse out.
By selecting an appropriate family of distributions for our EP approximation, we can efficiently integrate out parameters, leading to quicker mixing.
Importantly, through the use of EP, we still integrate over an approximation of our uncertainty when collapsing the nuisance variables.

Our experiments for the time series clustering model and mixture of Student-$t$ model demonstrate the effectiveness of our proposed approach.
More generally, we expect this approach to be useful in cases where collapsing involves a large number of latent variables. 

\presec
\section{Background}
\label{sec:background}

\pressec
\subsection{Latent Variable Models for Clustering}
\begin{figure}[ht!]
    \centering
   	\begin{minipage}[b]{.32\textwidth}
	\centering
		\includegraphics[height = .125\textheight]{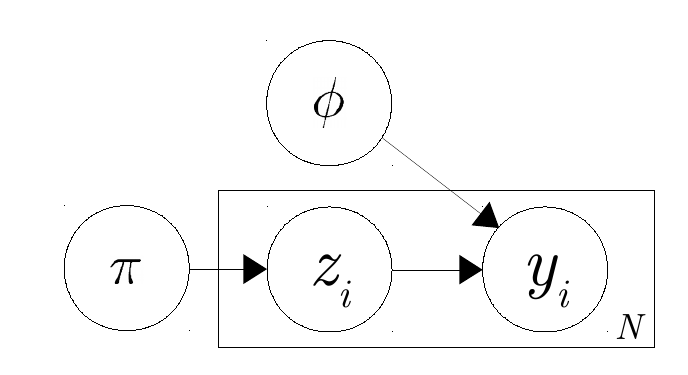}
    \end{minipage}
    \begin{minipage}[b]{.32\textwidth}
    \hspace{0.1in}
	\centering
		\includegraphics[height = .15\textheight]{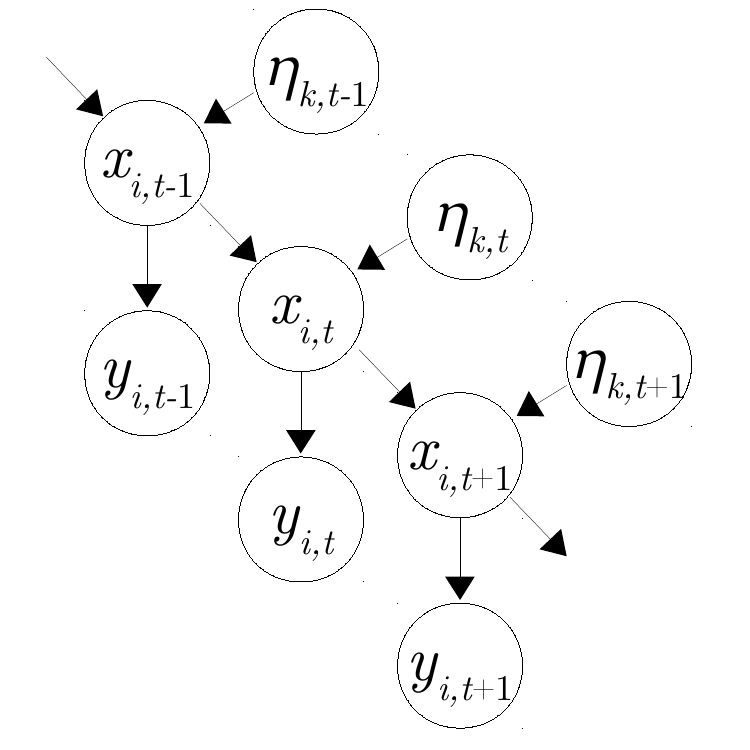}
    \end{minipage}
    \begin{minipage}[b]{.32\textwidth}
    \hspace{-0.1in}
	\centering
		\includegraphics[height =.175\textheight]{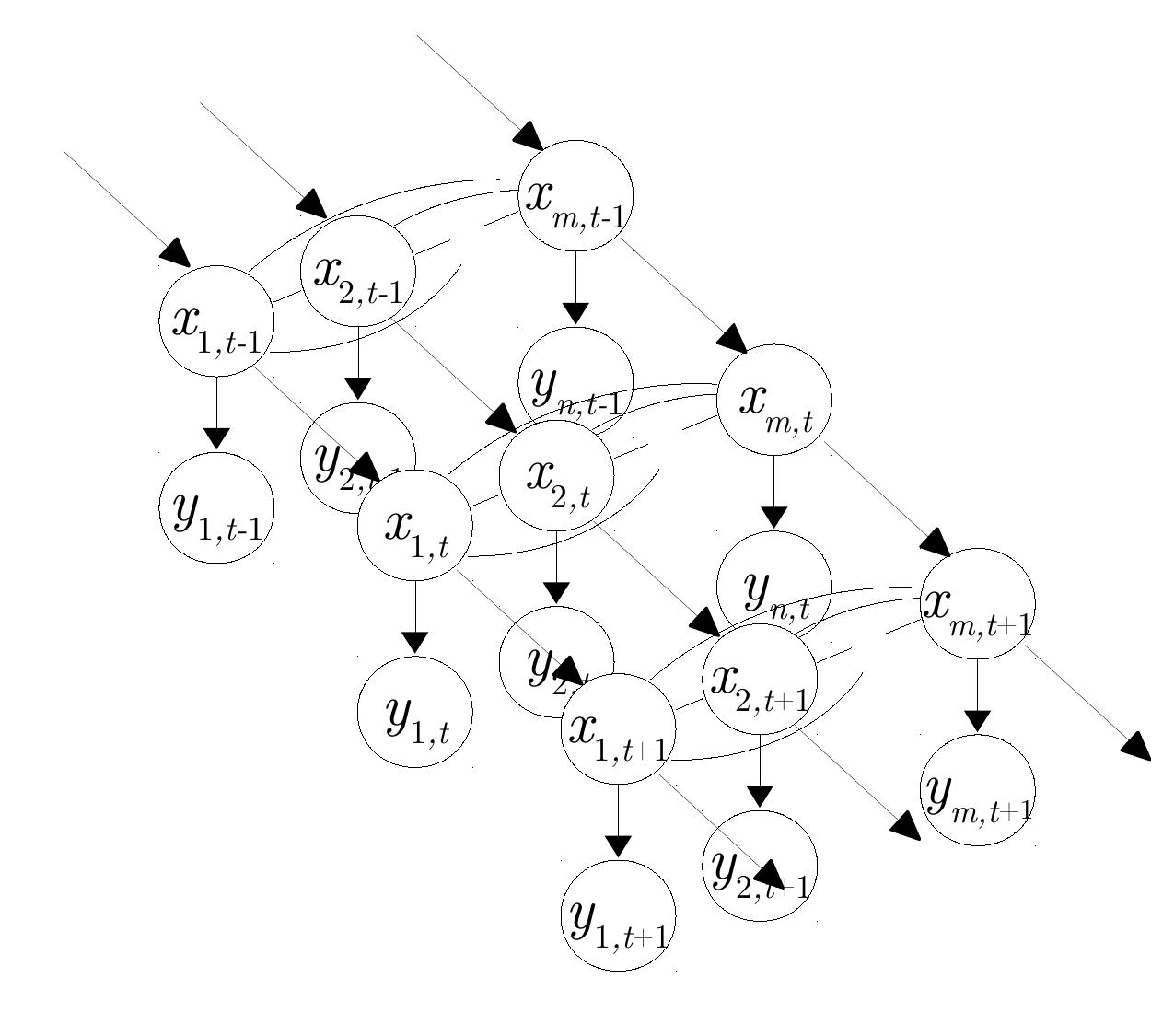}
    \end{minipage}
    %
    \caption{\small (Left) Generic clustering model.
    For the time series model of Sec.~\ref{sec:tscluster},
    individual time series likelihoods (center) without and (right) with collapsing the latent factor processes $\eta_{1:K,1:T}$.}
    \label{fig:GM}
\end{figure}

\postssec
\label{sec:background_model}
We first present the general abstract framework 
we assume when clustering data using latent
variable models.
We are interested in clustering $N$ observations $y = y_{1:N}$
into $K$ groups.
We assume that each observation $y_i$ has 
an associated latent variable $z_i \in \{1,\ldots,K\}$
denoting its cluster assignment.
We denote the cluster-specific parameters defining the observation distribution as $\phi = \phi_{1:K}$.
The distribution over the assignment variables is defined by parameters $\pi= \pi_{1:K}$.
Given $\phi,\pi$, 
\preeq
\begin{equation*}
    \Pr(y, z \mid \phi, \pi) = \prod_{i = 1}^N p_\phi(y_i \ | \ z_i,
    \phi)p_\pi(z_i \ | \ \pi) \enspace.
\end{equation*}
\posteq
The form of $p_\phi, p_\pi$ and the domains of $\phi,\pi$ depend on the application.
A Bayesian approach then specifies priors on $\pi,\phi$.
The generative process can be visualized as a graphical model in Fig.~\ref{fig:GM}(left).


\pressec
\subsection{Gibbs Sampling}
\postssec
The classic sampling approach for Bayesian inference in
the latent variable model of Sec.~\ref{sec:background_model}
is Gibbs sampling,
which (eventually) draws from the posterior 
by iteratively sampling from full conditionals. 

\prepar
\paragraph{Naive Gibbs Sampling}
The naive Gibbs sampler targets $\Pr(z, \pi, \phi \, | \, y)$ and iteratively samples each variable from
the posterior conditioned on the current value of \emph{all} other variables:

\begin{itemize}
    \item Sample $z^{(s+1)}_i \sim \Pr(z_i \ | \ y, z^{(s)}_{-i},
        \pi^{(s)}, \phi^{(s)})$ for all $i$
    \item Sample 
        $\pi^{(s+1)}, \phi^{(s+1)} \sim \Pr(\pi, \phi \ | \ y, z^{(s+1)})$
\end{itemize}
Here, $-i$ denotes all elements except $i$. The full conditional of $z_i$ decomposes into
the product of a prior and likelihood term
\preeq
\vspace{-0.1in}
\begin{equation}
\label{eq:naive_gibbs_factor}
    \Pr(z_i \ | \ y, z_{-i}, \pi, \phi) \propto
    p_\pi(z_i \ | \ \pi)
    \cdot
    p_\phi(y_i \ | \ z_i, \phi)
    \enspace.
\end{equation}
\posteq
Because we condition on the parameters $\phi, \pi$,
the observation $y_i$ and assignment
$z_i$ are conditionally independent of $y_{-i}, z_{-i}$
(see Figure \ref{fig:GM}(top)). Therefore, in naive Gibbs
we sample $z^{(s+1)}_i$ by
simply taking the product of the prior $p_\pi$ and likelihood $p_\phi$
for each possible cluster assignment and then normalizing.
This computation can be distributed across $i$ in an embarrassingly parallel
manner. One drawback of this naive Gibbs sampling scheme is that it can
\emph{mix} (i.e. move between regions of the posterior) extremely slowly.
This also impacts the speed at which we escape from poor initializations.

\prepar
\paragraph{Collapsed Gibbs Sampling}
To improve the mixing of naive Gibbs sampling,
collapsed Gibbs targets $\Pr(z \, | \, y)$
integrating out $\pi, \phi$ and then iterates

\begin{itemize}
\item Sample $z^{(s+1)}_i \sim \Pr(z_i \ | \ y, z^{(s)}_{-i})$ for all $i$
\end{itemize}
Similar to Eq.~\eqref{eq:naive_gibbs_factor} for naive Gibbs,
the conditional posterior can be decomposed into
the product of a prior and likelihood term
\preeq
\begin{equation}
\label{eq:collapsed_gibbs_factor}
    \Pr(z_i \ | \ y, z_{-i}) \propto
    \Pr(z_i \ | \ z_{-i}) \cdot
    \Pr(y_i \ | \ y_{-i}, z) \enspace.
\end{equation}
\posteq
In contrast to naive Gibbs, here things do not decouple across $i$ as
dependencies are introduced in the marginalization of $\pi, \phi$:
\preeq
\vspace{-0.1in}
\begin{align}
\label{eq:z_prior_int}
\Pr(z_i  |  z_{-i}) &=
    \int p_\pi(z_i  |  \pi) \Pr(\pi  |  z_{-i}) \ d\pi \\
\label{eq:z_like_int}
\Pr(y_i  |  y_{-i}, z) &= 
    \int p_\phi(y_i  |  z_i, \phi) \Pr(\phi  |  y_{-i}, z_{-i}) \
    d\phi \, .
\end{align}
\posteq
When the integrals of Eqs. \eqref{eq:z_prior_int} and \eqref{eq:z_like_int}
are tractable (e.g. due to conjugacy),
sampling $z$ from a collapsed Gibbs sampler can be considered. 
However when either of the integrals is intractable,
we cannot fully perform collapsed Gibbs sampling.
In practice, we integrate (or \emph{collapse}) out 
the variables that are analytically tractable and 
condition on those that are not~\cite{van2008partially, liu2008monte}.

\presec
\section{Approximate Collapsed Gibbs Sampling}
\postsec
\label{sec:inference}

Our goal is to develop efficient approximate collapsed Gibbs samplers
when the required integrals,
Eqs.~\eqref{eq:z_prior_int} and \eqref{eq:z_like_int}, are intractable.
Generically, we can write the intractable integrals of interest as
\begin{equation}
\label{eq:ep_target}
F(\zeta_i) = \int f_i(\zeta_i, \theta) p(\theta \, | \,  \zeta_{-i}) \ d\theta \enspace.
\end{equation}
where for Eqs.~\eqref{eq:z_prior_int} and~\eqref{eq:z_like_int}
\begin{align*}
    f_i(z_i, \theta) := p_\pi(z_i | \pi)\enspace &, \enspace
    p(\theta | z_{-i}) := \Pr(\pi | z_{-i}) \enspace, \\
    f_i(z_i, \theta) := p_\pi(y_i | z_i, \phi) \enspace &, \enspace
    p(\theta | z_{-i}) := \Pr(\phi | y_{-i}, z_{-i}) \enspace.
\end{align*}

We assume that Eq.~\eqref{eq:ep_target} is intractable
as $p(\theta | \zeta_{-i})$ either does not have an analytic form or
is computationally intractable to calculate. 
Because $p(\theta | \zeta_{-i})$ is intractable,
integral approximation methods, such as the Laplace approximation,
cannot be immediately applied to Eq.~\eqref{eq:ep_target}.

Our key idea is to replace $p(\theta | \zeta_{-i})$
with an approximate distribution $q(\theta | \zeta_{-i})$ such that
\begin{equation}
\label{eq:ep_approx}
\tilde{F}(\zeta_i) = \int f_i(\zeta_i, \theta) q(\theta \ | \ \zeta_{-i}) \ d\theta \enspace
\end{equation}
is a good approximation to $F(\zeta_i)$ in Eq.~\eqref{eq:ep_target}.
To do this, we borrow ideas from EP, an iterative method for minimizing the
Kullback-Leibler (KL) divergence between
an approximation $q$ and a posterior $p$.

\pressec
\subsection{Review of Expectation Propagation}
\postssec
\label{sec:EPreview}
We briefly review EP before describing how we use these ideas to form the approximation in Eq.~\eqref{eq:ep_approx}.
Traditionally, EP has been used to approximate
the posterior for some $\theta$ given all observations $y$
\begin{equation}
p(\theta | y) \propto p_0(\theta) \cdot \prod_{j = 1}^N f_j(y_j, \theta)
    \enspace,
\end{equation}
where $p_0(\theta)$ is the prior for $\theta$. The EP idea is to approximate the likelihood terms $f_j(y_j, \theta)$ with 
\emph{site approximations} $\tilde{f}_j(\theta)$ that
are conjugate to the prior.
For example, if the prior $p_0(\theta)$ is Gaussian, then
\begin{equation}
    \tilde{f}_j \in \tilde{\mathcal{F}} = \{ C_j \mathcal{N}(\theta | \mu_j,
    \Sigma_j) \, : \, C_j > 0, \mu_j \in \R^n, \Sigma_j \in \mathbb{S}^{n} \}
\end{equation}
where $\mathcal{N}(\cdot | \mu, \Sigma)$ denotes
a multivariate Gaussian density with mean $\mu$ and variance
$\Sigma$ and $C$ is a scaling constant.
Note that $\tilde{f}_j$ is a \emph{likelihood approximation},
not a probability density,
thus its parameterization does not necessarily integrate to one.
See~\cite{seeger2005expectation}.

The resulting approximation $q(\theta | y)$ is then
\vspace{-0.1in}
\begin{equation}
q(\theta | y) \propto p_0(\theta) \cdot \prod_{j = 1}^N \tilde{f}_j(\theta)
    \enspace.
\label{eq:EPapprox}
\end{equation}
\vspace{-0.1in}

To construct good site approximations $\tilde{f}_{1:N}$ for $f_{1:N}$,
EP attempts to minimize $\text{KL}(p(\theta | y) || q(\theta | y))$. 
Directly minimizing this KL divergence is intractable due to the integral
with respect to $p(\theta | y)$.
Instead, EP iteratively selects each $\tilde{f}_i$ to minimize a local KL
divergence~\cite{minka2001expectation}:
\begin{equation}
\label{eq:ep_update}
\tilde{f}_i(\theta) =
\argmin_{\tilde{f} \in \tilde{\mathcal{F}}} 
    \text{KL}\left( f_i(y_i, \theta) q(\theta | y_{-i})  \ || \
                    \tilde{f}(\theta) q(\theta | y_{-i})) \right) \enspace.
\end{equation}
Here, $q(\theta | y_{-i})$ is the \emph{cavity distribution} for site $i$, which take the form of Eq.~\eqref{eq:EPapprox} with $\tilde{f}_i(\theta)$ removed.

Minimizing each local KL divergence, Eq.~\eqref{eq:ep_update},
is equivalent to matching $\theta$'s sufficient statistics' moments.
This can be done analytically for a wide class of
distributions~\cite{seeger2005expectation}.
EP iteratively updates $\tilde{f}_i$ until convergence,
which can be ensured by damping
~\cite{heskes2002expectation, teh2015distributed}.

\pressec
\subsection{EP for Approximate Collapsed Gibbs}
\postssec

There are a couple of necessary leaps to see how we apply EP to approximate the integrals of Eqs.~\eqref{eq:z_prior_int} and \eqref{eq:z_like_int}.
First, instead of approximating the posterior $p(\theta | y)$ with $q(\theta | y)$,
we are interested in approximating $p(\theta | \zeta_{-i})$ with the cavity distribution $q(\theta | \zeta_{-i})$ for each $i$.
Recall from Eq.~\eqref{eq:ep_target} that $\zeta_i$ can consist of just $z_i$ or $y_i$ as well;
neither of these is typically targeted by EP.
%
%

Note that our target distribution $p(\theta | \zeta_i)$
is conditioned on the \emph{sampled} latent variables $z_{-i}$.
In contrast to a fixed target distribution $p(\theta | y)$,
our target distribution $p(\theta | \zeta^{(s)}_{-i})$ changes
as we sample $z^{(s)}$ at every iteration $s$.
Therefore, the fixed points of our update scheme, the best EP approximation
$q^*(\theta | \zeta^{(s)})$,
change at each iteration.
To ensure stable performance,
one approach would be to run EP to convergence at every iteration

\preitem
\begin{itemize}
    \item Sample $z^{(s+1)}_i \sim \tilde{\Pr}(z_i \, | \, y, z_{-i}^{(s)})$
        approximately integrating out $\pi, \phi$ using
        $q^{*}(\theta \, | \, \zeta_{-i}^{(s)})$.
    \item Calculate $q^{*}(\theta \, | \, \zeta^{(s+1)})$ by updating
        \emph{all site} approximations $\tilde{f}_j(\theta)$ until convergence.
\end{itemize}
\postitem
At each iteration, at most one latent variable $z_i$ changes; therefore
we only need to update the site approximations $\tilde{f}_j(\theta)$
belonging in $z_i^{(s)}$ and $z_i^{(s+1)}$.
However, this is computationally costly 
as updating all sites in both clusters would take O($N$) time.

Instead, we choose a second approach that leverages our existing approximation 
$q^{(s)}(\theta | \zeta^{(s)})$
and only updates site approximation $\tilde{f}_i$ after sampling $z_i$.
That is,

\preitem
\begin{itemize}
    \item Sample $z^{(s+1)}_i \sim \tilde{\Pr}(z_i \, | \, y, z_{-i}^{(s)})$
        approximately integrating out $\pi, \phi$ using
        $q^{(s)}(\theta \, | \, \zeta_{-i}^{(s)})$.
    \item Calculate $q^{(s+1)}(\theta \, | \, \zeta^{(s+1)})$ by updating
        \emph{only} site approximation $\tilde{f}_i(\theta)$.
    \item Periodically (e.g. after a full pass),
        update all site approximations until convergence.
\end{itemize}
\postitem

By not updating all site approximations to convergence,
we introduce some error between
our new approximation $q^{(s)}(\theta | \zeta^{(s)})$ and 
the best EP approximation $q^*(\theta | \zeta^{(s)})$.
This error arises due to using `stale' site approximations.
The idea is similar in spirit to ``Parameter Server'' methods
that infer global parameters by passing `stale' sufficient statistics
between machines~\cite{li2014scaling,ahmed2012scalable}.
Intuitively, we expect this error to be small when our approximating family
closely resembles the likelihood and 
when the latent variables $z$ change slowly.
By periodically including full EP update passes,
we can bound the convergence error between this sparse update scheme and full
EP (up to model specific constants).
A more precise description and analysis of this convergence error,
including illustrative synthetic experiments,
can be found in the Appendix~\ref{app:ep_converge}. 
In our experiments (Sec.~\ref{sec:experiments}),
we found that it was even sufficient to omit the full pass and only update
the local site approximation at each iteration.
%

\presec
\section{Case Studies}
\postsec
\label{sec:case_studies}
We consider two motivating examples for the use of our EP-based approximate
collapsed Gibbs algorithm.
The first is a mixture of Student-$t$ distributions, which can capture heavy-tailed
emissions crucial in robust modeling (i.e., reducing sensitivity to outliers).
The second example is a time series clustering model.


\pressec
\subsection{Mixture of Multivariate Student-$t$}
\postssec
\label{sec:student}
The multivariate Student-$t$ (MVT) distribution, is a popular method
for handling robustness \cite{portilla2003image,peel2000robust,bishop2004robust,andrews1974scale}.
To perform robust Bayesian clustering of data $y = y_{1:N}$ in $\R^d$,
we use MVT as the emission distributions:
\begin{align}
    \label{eq:student_likelihood}
    &p_\phi(y_i \, | \, z_i = k, \mu, \Sigma, \nu) := 
        t_{\nu_k}(y_i \, | \, \mu_k, \Sigma_k) \enspace, \\
    &t_\nu(y \, | \, \mu, \Sigma) \propto |\Sigma|^{-1/2}
        \left(1 + \frac{(y_i - \mu)^T\Sigma(y_i-\mu)}{\nu}
        \right)^{-\frac{\nu+d}{2}}
        , 
    \nonumber
\end{align}
where $\mu_k, \Sigma_k, \nu_k$ are the mean, covariance matrix and degrees of freedom parameter for cluster $k$, respectively.
A common construction of the MVT arises from a scale mixture of Gaussians
\begin{equation}
    t_\nu(y \, | \, \mu, \Sigma) = \int \mathcal{N}(y \, | \, \mu, \Sigma/u) \Gamma(u \, | \, \nu/2, \nu/2) \, du \enspace.
\label{eq:student_scalemix}
\end{equation}

For this paper, we focus on the case where $\nu$ is known and
learn $z, \mu, \Sigma$ by Gibbs sampling.
However, all of the following sampling strategies can be
extended to learn $\nu$ by adding a Metropolis-Hasting step.



\prepar
\paragraph{`Naive' and Collapsed Gibbs:}
Because the MVT likelihood is non-conjugate to
standard exponential family priors,
the posterior conditional distribution for $\mu, \Sigma$ does not have 
a closed analytic form.
However, by exploiting the representation of a MVT 
as a scale mixture of Gaussians, Eq.~\eqref{eq:student_scalemix},
we can use data augmentation to construct a Gibbs sampler with analytic steps~\cite{liu1996bayesian, damlen1999gibbs}.
By introducing auxiliary variables $u = \{u_{i,k}\}$ 
for each observation-cluster pair $i,k$,
we can replace the MVT likelihood with a Gaussian conditioned on $u$.

The \texttt{naive Gibbs} sampler can be straightforwardly derived 
on the expanded space of $z, \mu, \Sigma, u$ as
\begin{align*}
&    z_i \, | \, z_{-i}, \mu, \Sigma, u, y \sim \mathcal{N}(y_i \, | \, \mu_{z_i}, \Sigma_{z_i}/u_{i, z_i}) \Pr(z_i \, | \, z_{-i}) \\
&    \mu_k, \Sigma_k \, | \, z, u, y \sim 
    \mathcal{NIW}(\mu, \Sigma \, | \, \tau_{\mu_k,\Sigma_k}(z,u,y))\\
&     u_{i,k} \, | \, z, \mu, \Sigma, y \sim \Gamma(u \, | \, \tau_{u_{i,k}}(z,\mu, \Sigma,y))
\enspace,
\end{align*}
where the specific form of the parameters
$\tau$ is given in the Appendix~\ref{app:student}.

Conditioned on $u$,
the posterior for $\mu, \Sigma$ is conjugate 
to the likelihood of $y$.
Therefore, we can integrate out $\mu, \Sigma$ when sampling $z$.
We cannot completely collapse out $\mu, \Sigma$ as
they are required for sampling $u$ (which is required for sampling $z$).
The result is a (partially collapsed) \texttt{blocked Gibbs} sampler
that samples
\begin{align*}
&\begin{rcases}
&   z_i \, | \, z_{-i}, u, y \sim t(y_i \, | \, \tau(z, u)) \Pr(z_i \, | \, z_{-i}) \\
&   \mu_k, \Sigma_k \, | \, z, u, y \sim 
    \mathcal{NIW}(\mu, \Sigma \, | \, \tau_{\mu_k,\Sigma_k}(z,u,y))
\end{rcases} \text{ block } \\
& \ \quad u_{i,k} \, | \, z, \mu, \Sigma, y \sim \Gamma(u \, | \, \tau_{u_{i,k}}(z,\mu, \Sigma,y))
\enspace.
\end{align*}

For further details, see the Appendix~\ref{app:student}.

Although the data-augmentation method allows us to construct
analytic Gibbs samplers for a mixture of MVTs,
this approach has serious drawbacks.
By expanding the representation of the MVT with $u$,
we (i) increase the number local modes and
(ii) increase computation by sampling $NK$ auxiliary parameters.
For these reasons, the data-augmentation approach is not commonly used beyond small $K$.

\prepar
\paragraph{Approximate Collapsed Gibbs:}
We can handle the non-conjugacy of the MVT likelihood 
(Eq.~\eqref{eq:student_likelihood})
using our framework by approximately collapsing out $\mu, \Sigma$ 
without introducing auxiliary variables $u$.

%

Using the Gaussian scale mixture representation of the MVT, 
the collapsed likelihood term for $z_i$ is
\begin{equation}
    \label{eq:student_collapsed}
    F(\zeta_i) = \int 
    \underbrace{\Pr(\mu_{z_i}, \Sigma_{z_i} \, | \, y_{-i}, z_{-i})
    }_{p(\theta|\zeta_{-i})} \times
    \underbrace{
    \int \mathcal{N}(y_i \, | \, \mu_{z_i}, \Sigma_{z_i}/u) 
    \Gamma(u | \nu/2, \nu/2) \ du}_{f_i(\zeta_i, \theta)}
    \, d\mu \, d\Sigma
\end{equation}
By selecting our approximation family $q(\mu_k, \Sigma_k|\zeta)$ to be a $D$-dimensional
normal inverse-Wishart, and by swapping the order of integration,
our approximation to Eq.~\eqref{eq:student_collapsed}
becomes a 1-dimensional integral over $u$
\begin{equation}
    \label{eq:student_ep_approx}
    \tilde{F}(\zeta_i) = \int  \Gamma(u | \nu/2, \nu/2) \times
    \underbrace{\int q(\mu_{z_i}, \Sigma_{z_i}|\zeta_{-i})
    \mathcal{N}(y_i \, | \, \mu_{z_i}, \Sigma_{z_i}/u)
    \, d\mu \, d\Sigma
    }_{\propto \text{ normal inverse-Wishart}(u)}
     \, du
\end{equation}
Because the integrand is a ratio of normal inverse-Wishart normalizing constants,
which are analytically known, this 1-dimensional integral can be calculated numerically.
Similary, the moments required for EP are also calculated as a
1-dimensional integral of weighted normal inverse-Wishart sufficient statistics.
For complete details, see the Appendix~\ref{app:student}.

For the MVT, the key innovation is using EP to keep track of an approximation $q(\theta \, | \, \zeta_{-i})$ (here a normal inverse-Wishart) for $p(\theta \, | \, \zeta_{-i})$,
thus allowing Eq.~\eqref{eq:student_ep_approx} to be numerically tractable.
This approach allows us to approximately collapse out $\mu, \Sigma$,
which in turn enables us to avoid sampling the auxiliary variables $u$ introduced in the naive and blocked samplers.

\pressec
\subsection{Time Series Clustering}
\postssec
\label{sec:tscluster}

\commentout{
\begin{figure*}
    \vskip 0.1in
    \centering
    \begin{minipage}[b]{.45\textwidth}
	\centering
		\includegraphics[height = .2\textheight]{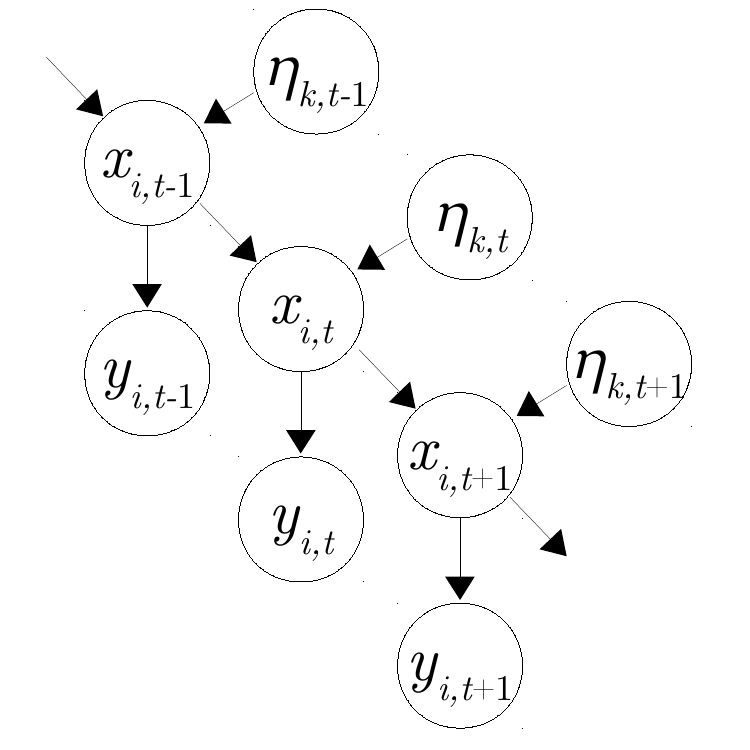}
    \end{minipage}
    \begin{minipage}[b]{.45\textwidth}
	\centering
		\includegraphics[height = .2\textheight]{./figures/GM_collapsed_ts2.pdf}
    \end{minipage}
    \caption{(Left) graphical model for an individual series likelihood,
    (Right) graphical model of likelihood after collapsing out $\eta$.}
    \label{fig:GM_ts}
    \vskip -0.1in
\end{figure*}
}

Given a collection of time series, we are interested in finding clusters of series
such that series within a cluster are \emph{correlated} and between 
clusters are \emph{independent}.
We take motivation from a housing application analyzed by
\citet{ren2015achieving}.
The goal is to estimate housing price trends at fine spatial resolutions.
The series cannot be analyzed independently 
while providing reasonable value estimates 
due to the scarcity of spatiotemporally localized house sales observations.
The time series clustering helps handle this data scarcity by sharing information across regions discovered to be related.
%


Let $y = \{y_i \in \R^T \}_{i = 1}^N$ be a collection of $N$ observed
time series of length $T$ (different lengths and missing data can also be
accommodated).
The individual series follow a state space model:
\begin{align}
    x_{i, t} &= a_i x_{i, t-1} + \epsilon_{i, t} &
    \epsilon_{i,t} \sim \mathcal{N}(0,\sigma_{i,t}^2) \nonumber\\
    y_{i, t} &= x_{i, t} + \nu_{i, t} &
    \nu_{i,t} \sim \mathcal{N}(0, \sigma_{y_i}^2)
    \enspace.
    \label{eq:general_correlated_model}
\end{align}
Here, $x_{i,t} \in \R$. Clusters of correlated time series are induced by introducing latent
cluster assignments $z$ and taking $\epsilon_{i,t}$ to follow a cluster-specific latent factor process $\eta_{z_i, t}$ with factor loading $\lambda_i$:
\begin{align}
    \epsilon_{i,t} &= \lambda_i \eta_{z_i, t} + \tilde{\epsilon}_{i,t} &
    \eta_{k,t} \sim \mathcal{N}(0,1)\, , 
    \tilde{\epsilon}_{i,t} \sim \mathcal{N}(0, \sigma_x^2)
    \enspace.
    \label{eq:correlated_noise_process}
\end{align}
Marginalizing over $\eta_{k,t}$, $\CoV(\epsilon_{i,t}, \epsilon_{j,t} | z) = \lambda_i \lambda_j + \sigma^2_x \mathbf{1}_{i = j}$ if $z_i = z_j = k$ and 0 otherwise. 
%
Combining Eq.~\eqref{eq:general_correlated_model} and
\eqref{eq:correlated_noise_process},
an equivalent representation for the individual latent series dynamics is
\begin{align}
    x_{i,t} &= a_i x_{i, t-1} + \lambda_i \eta_{z_i, t} + \tilde{\epsilon}_{i, t} \enspace.
    \label{eq:eta_model}
\end{align}



\prepar
\paragraph{`Naive' and Collapsed Gibbs:}
The simplest Gibbs sampler is to iteratively sample all variables, including the latent states $x$.
Instead, as in \citet{ren2015achieving},
we exploit the time series structure of the state space model and
always integrate out $x$ using a Kalman smoother~\cite{barber2011inference,
bishop2006pattern} with a slight modification to account for the time-varying mean term
$\lambda_i\eta_{z_i}$:
\begin{equation}
    \label{eq:ts_naive_z_like}
    p_\phi(y_i  | z_i, \phi) := \Pr(y_i  |  z_i, \eta) 
    = \int \prod_{t = 1}^T \Pr(y_{i,t}  |  x_{i,t})
    \Pr(x_{i,t}  |  x_{i,t-1}, \eta_{z_i,t}) \ d x_t \enspace.
\end{equation}
See Fig.~\ref{fig:GM}(center) for a visualization of this partially collapsed
likelihood.
In this model, the Dirichlet prior $p_\pi$ over cluster assignments $z$ is
conjugate and can be analytically marginalized.
We refer to this partially collapsed scenario that conditions on the latent
factor processes $\eta_{1:K,1:T}$ as \texttt{naive Gibbs}.
%
Note that running the Kalman smoother on one series has a runtime complexity of 
$O(T)$~\cite{barber2011inference}. 
By evaluating this for each potential cluster assignment,
sampling $z_i$ has a total runtime complexity of $O(TK)$.
Unfortunately, this naive sampler is sensitive to initialization and 
exhibits poor performance.

To overcome this, \citet{ren2015achieving}
constructed a collapsed Gibbs sampler that additionally integrates out $\eta$.
From the state space model of Eq.~\eqref{eq:eta_model},
collapsing out $\eta$ induces dependencies between the latent states $x$
assigned to the same cluster
(see Fig.~\ref{fig:GM}(right)).
The conditional covariance structure is specified under Eq.~\eqref{eq:correlated_noise_process}. 
As a result, calculating the collapsed likelihood term requires
running the Kalman smoother on all series $y_j$ assigned to the same cluster.
Although analytically tractable,
the computational complexity of the Kalman smoother scales cubically
in the number of series~\cite{barber2011inference}.
Therefore, the collapsed likelihood is computationally prohibitive
for large cluster sizes.
We refer to this sampler as \texttt{collapsed Gibbs}.


\prepar
\paragraph{Approximate Collapsed Gibbs:}
We apply our framework of Sec.~\ref{sec:inference} to reduce
the computational overhead by 
calculating an approximation to the collapsed likelihood term
\begin{equation}
\label{eq:ts_target}
    F(\zeta_i) = \int \underbrace{\Pr(\eta \ | \ y_{-i}, z_{-i})}_{
        p(\theta | \zeta_{-i}) } \times 
    \underbrace{\prod_{t = 1}^T\Pr(y_{i,t} \ | \ x_{i,t})
        \Pr(x_{i,t} \ | \ x_{i,t-1}, \eta_{z_i,t})  \ d x_i}_{
            f_i(\zeta_i, \theta)}\ d\eta \enspace.
\end{equation}

By selecting $q(\eta_k \ | \ z) \in \{ \mathcal{N}_T(\eta_k \ | \ \mu_k, \diag(\sigma_k))\}$, i.e., as a $T$-dimensional diagonal Gaussian, 
we can factorize $q$ over $t$ and approximate Eq.~\eqref{eq:ts_target} with
\begin{equation}
\tilde{F}(\zeta_i) = 
    \int \prod_{t = 1}^T q(\eta_{z_i, t} \ | \ \zeta_{-i}) \times 
    \Pr(y_{i,t} \ | \ x_{i,t}) 
    \Pr(x_{i,t} \ | \ x_{i,t-1}, \eta_{z_i,t}) 
      \ d x_i \ d\eta \enspace.
\end{equation}
This integrand has the same graphical model form as in the naive Gibbs case
(Fig.~\ref{fig:GM}(center)) and can be calculated in $O(T)$ time using the
Kalman smoother modified to account for $\eta$.

To update the site approximations $\tilde{f}_i(\eta) \in \{ C_i \mathcal{N}_T(\eta \ | \ \mu_i, \diag(\sigma_i))\}$,
we must calculate the marginal mean and variance of $\eta_{k,t}$.
Fortunately, the moments of $\eta_{k,t}$ can be calculated given the
pairwise distribution of $(x_{t}, x_{t+1})$ extracted from
the Kalman smoother.
For further details, see the Appendix~\ref{app:tscluster}.


\presec
\section{Experiments}
\postsec
\label{sec:experiments}

To assess the computational complexity and cluster assignment mixing of our
sampling methods, we perform experiments on both synthetic and real data from
the considered models of Secs.~\ref{sec:student} and \ref{sec:tscluster}.

To evaluate our sampling methods,
we measure the normalized mutual information (NMI) of 
the inferred cluster assignment to the ground truth when known.
When the clustering is not known,
we compare to the clustering associated with the MAP of the exact collapsed sampler run for a long time.
NMI is an information theoretic measure of similarity between cluster assignments~\cite{vinh2010information}.
NMI is maximized at 1 when the assignments are equal up to a permutation and minimized at 0 when the assignments share no information.

\presec
\subsection{Mixture of Multivariate Student-$t$}
\postsec

%

\begin{figure}[htb]
    \centering
    \begin{minipage}[t]{.45\textwidth}
    \centering
        \includegraphics[width=\textwidth]{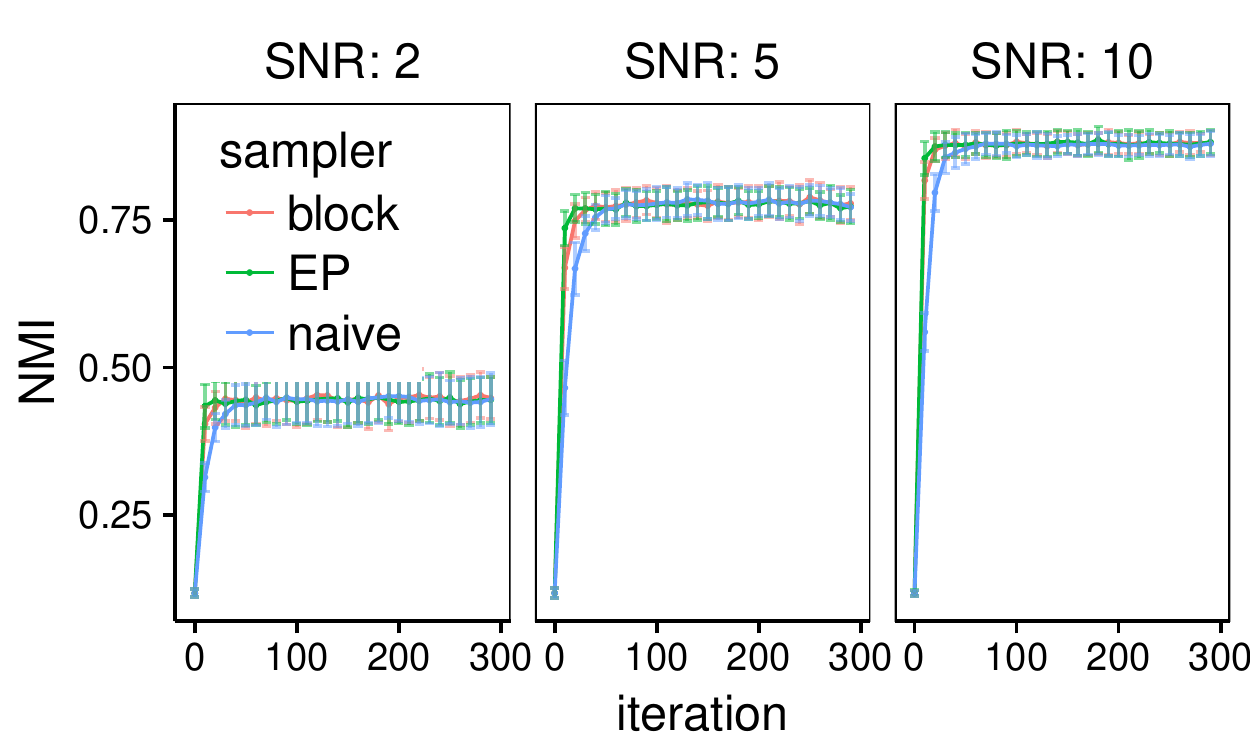}
    \end{minipage}
    \hspace{0.1in}
    \begin{minipage}[t]{.45\textwidth}
    \centering
        \includegraphics[width=\textwidth]{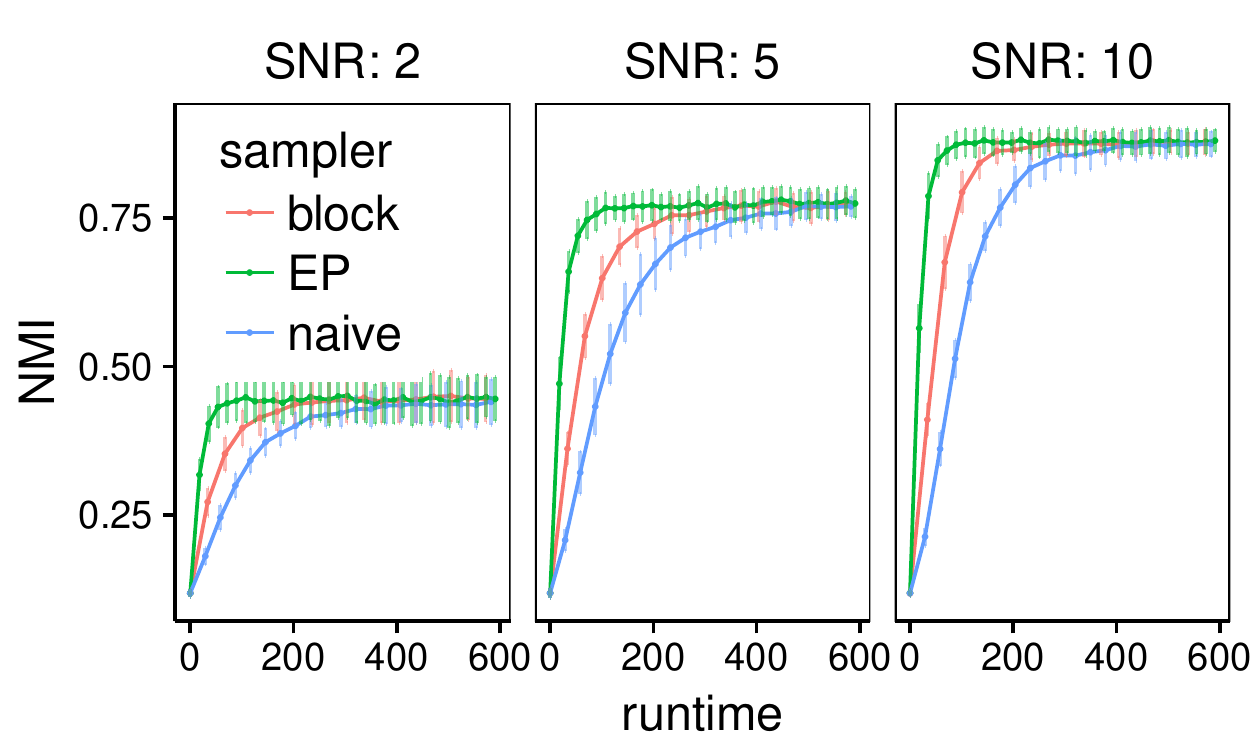}
    \end{minipage}
\vspace{-0.1in}
    \caption{\small Comparing three Gibbs samplers on synthetic data:
    \textcolor{BrickRed}{blocked} Gibbs,
    our \textcolor{ForestGreen}{EP} approximate collapsed Gibbs, and
    \textcolor{Blue}{naive} Gibbs.
    (Top) NMI vs iteration, (right) NMI vs runtime.
    The mean value and error bars are over 25 trials.}
    \label{fig:synth_student}
    \vskip -0.2in
\end{figure}

\begin{figure}[htb]
    \vskip 0.1in
    \centering
%
    %
    \begin{minipage}[t]{.29\textwidth}
    \centering
        \includegraphics[width=\textwidth]{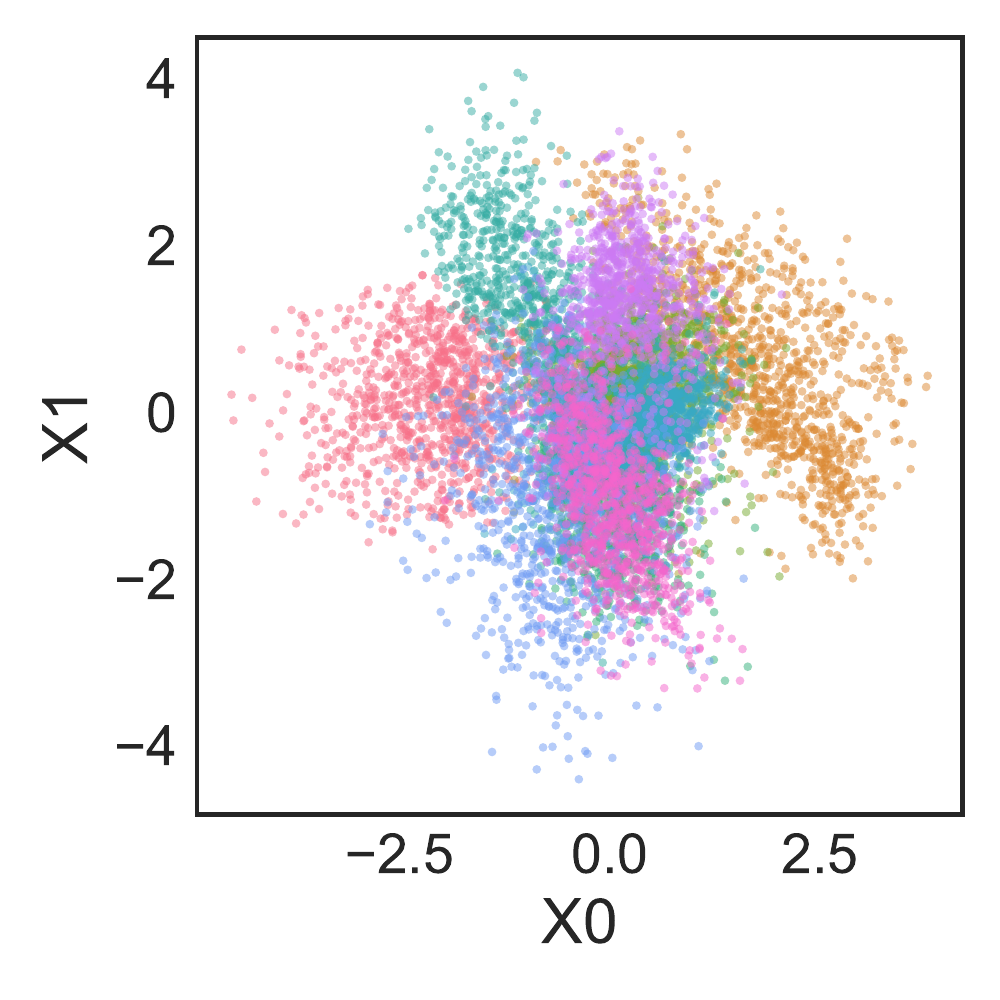}
    \end{minipage}
    \hspace{0.1in}
    \begin{minipage}[t]{.29\textwidth}
    \centering
        \includegraphics[width=\textwidth]{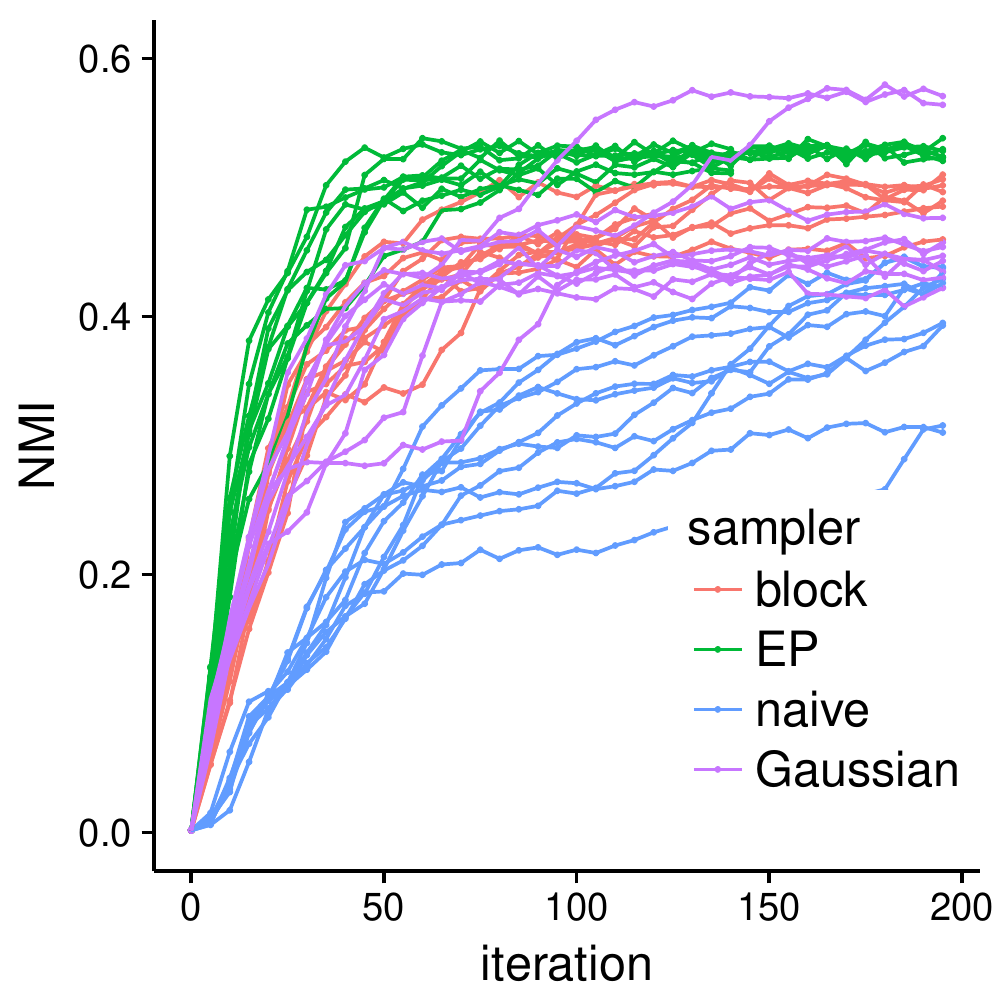}
    \end{minipage}
    \begin{minipage}[t]{.29\textwidth}
    \centering
        \includegraphics[width=\textwidth]{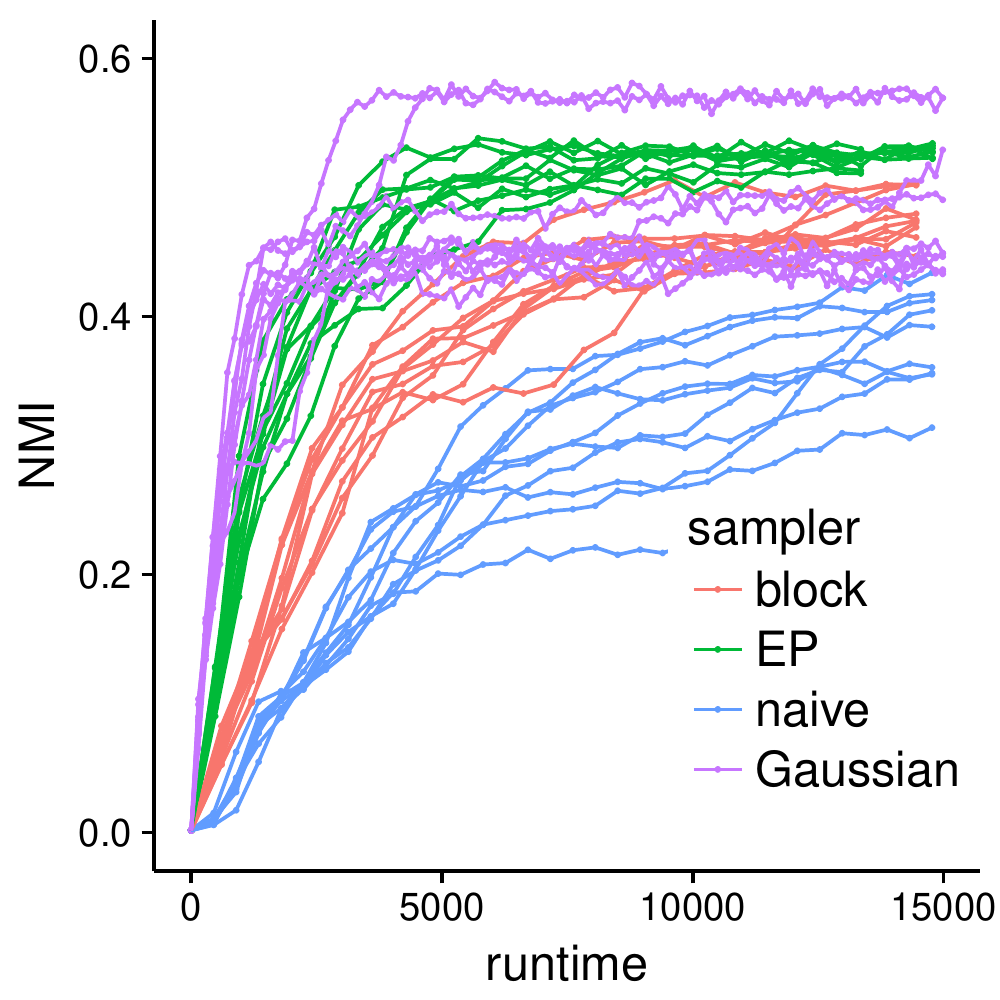}
    \end{minipage}
\vspace{-0.1in}
    \caption{\small Comparing four Gibbs samplers on the VAE embedding of \textit{MNIST}:
    \textcolor{BrickRed}{blocked} Gibbs,
    our \textcolor{ForestGreen}{EP} Gibbs,
    \textcolor{Blue}{naive} Gibbs,
    and \textcolor{Purple}{Gaussian} mixture model.
    (Left) VAE embedding, (middle) NMI vs iteration, (right) NMI vs runtime.
    }
    \label{fig:mnist}
    \vskip -0.2in
\end{figure}


We consider fitting mixtures of MVT to synthetic data and 
a low-dimensional variational auto-encoder embedding of the MNIST dataset.
We compare the \texttt{naive Gibbs}, \texttt{blocked Gibbs} and approximate collapsed \texttt{EP Gibbs} samplers
of Sec.~\ref{sec:student}.
For this model, the exact collapsed sampler is not available.


For our synthetic experiments, we generated data from a mixture of MVTs
with $\nu = 5$, $K=20$, and $N=600$.
The cluster mean and variance parameters $\mu, \Sigma$ were drawn from the normal inverse-Wishart.
We considered three different signal-to-noise (SNR) settings by increasing the variance of $\mu$ ranging from hard to easy.
Fig.~\ref{fig:synth_student} shows the performance of each sampler. 
From the iteration plot (top), we see that all methods have similar performance.
From the runtime plot (bottom), we see that \texttt{EP Gibbs} > \texttt{blocked Gibbs} > \texttt{naive Gibbs}.

For our real dataset example, we consider clustering a $\R^3$ embedding of MNIST handwritten digit images~\cite{lecunmnisthandwrittendigit2010},
where the ground truth cluster assignments are taken to be the true digit-labels.
A simple past approach to clustering MNIST consists of running PCA to learn a low dimensional embedding followed by clustering.
Instead of PCA, we use variational autoencoders (VAEs), an increasingly popular and flexible method for unsupervised learning of complex distributions~\cite{kingma2013auto}.
VAEs learn a probabilistic encoder to infer a latent embedding such that the latent embedding comes from a simple distribution (usually an isotropic Gaussian).
In practice, when we the data come from different classes, the VAE warps the clusters apart making them non-Gaussian.

We trained a simple VAE on the MNIST dataset with latent embedding dimension $3$ using the same architecture as in \citep{kingma2013auto}.
The scatter plot, Fig.~\ref{fig:mnist}(left), visualizes the VAE embedding, with separate colors for each digit.

We fit the MVT samplers from Sec~\ref{sec:student} using $\nu = 5$ and $K=10$ on a stratified subset of MNIST $(N=10000)$.
In addition, we also fit a \texttt{Gaussian} mixture model using a collapsed Gibbs sampler
to illustrate the potential advantage of the more robust MVT likelihood.
In Fig.~\ref{fig:mnist}, we present the results comparing each sampler's clustering assignment with the ground truth labels.
Fig.~\ref{fig:mnist}(middle) plots NMI vs iteration. 
We see that the MVT \texttt{EP Gibbs} and \texttt{blocked Gibbs} methods out perform the \texttt{Gaussian} mixture model per iteration (on average).
Fig.~\ref{fig:mnist}(right) is NMI vs runtime. We see that \texttt{EP Gibbs} is much faster than the alternative data-augmentation MVT samplers (due to sampling $u_{i,k}$).
We expect the runtime improvements of EP over data-augmentation Gibbs to be greater for larger $K$.

\pressec
\subsection{Time Series Clustering}
\label{sec:sea_housing_data}
\postssec
\begin{figure*}[htb]
    \centering
    \begin{minipage}[b]{.3\textwidth}
	\centering
		\includegraphics[width=\textwidth]{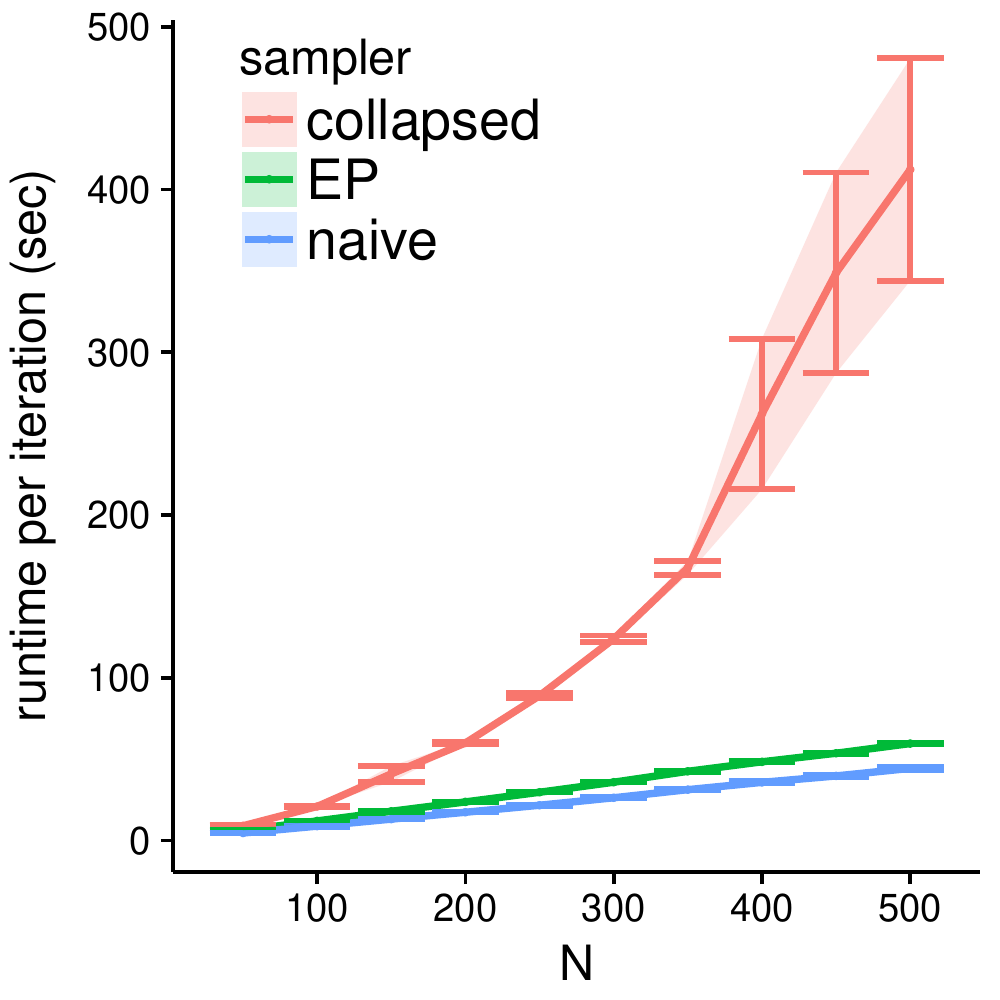}
    \end{minipage}
    \begin{minipage}[b]{.3\textwidth}
	\centering
		\includegraphics[width=\textwidth]{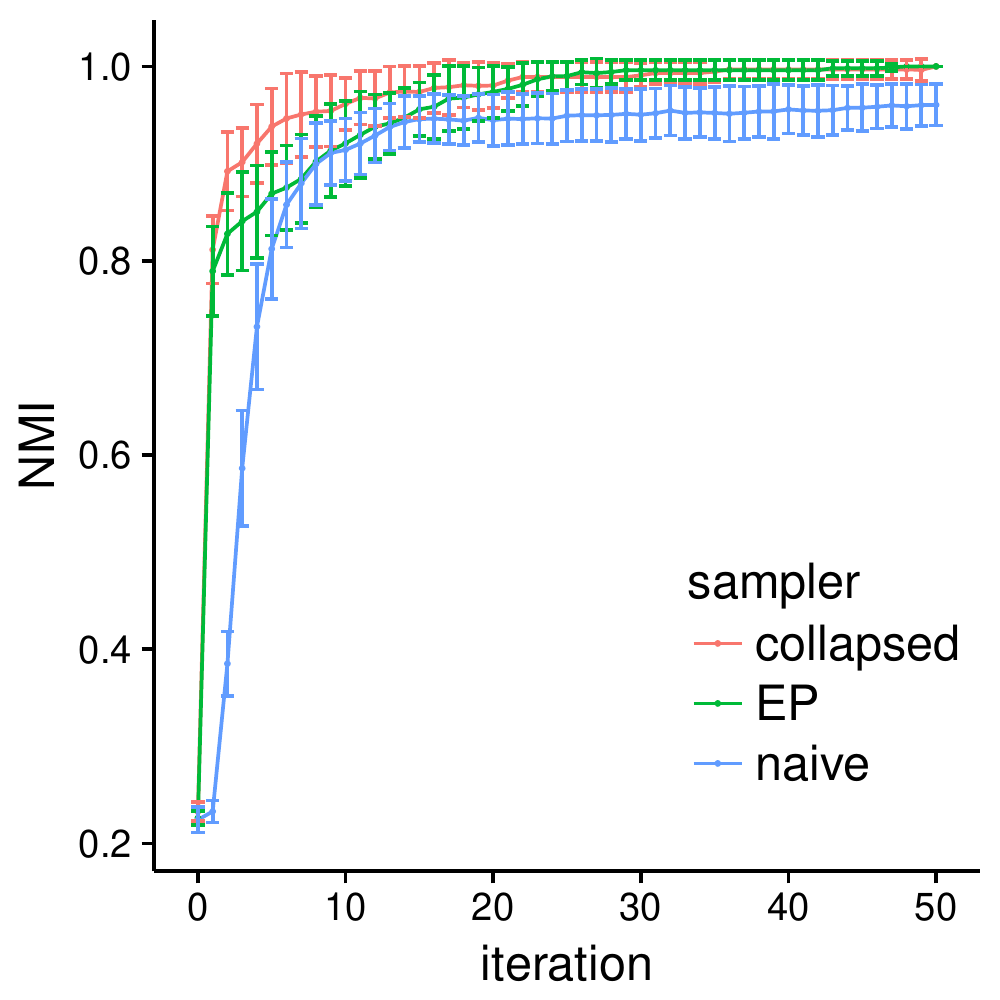}
    \end{minipage}
    \begin{minipage}[b]{.3\textwidth}
	\centering
		\includegraphics[width=\textwidth]{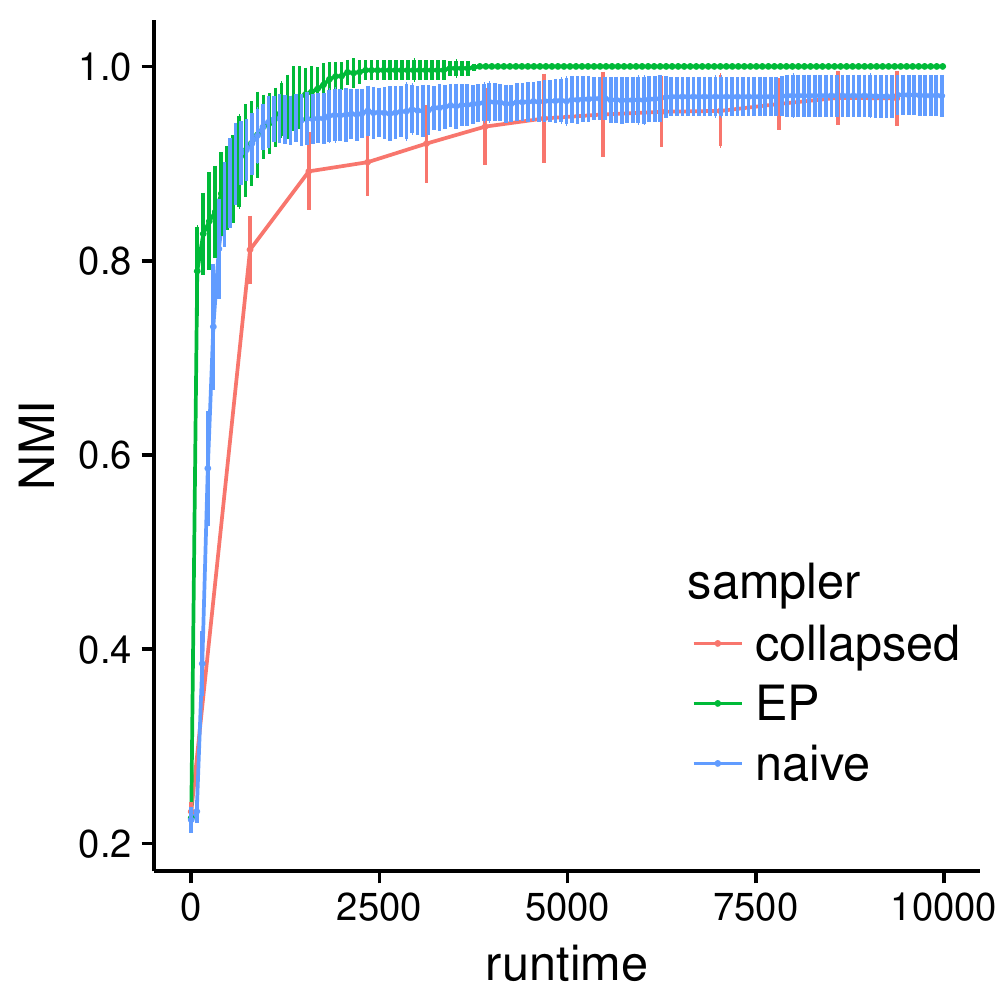}
    \end{minipage}
\vspace{-0.1in}
    \caption{\small Comparing three Gibbs
    samplers on synthetic time series:
    \textcolor{BrickRed}{collapsed} Gibbs,
    our \textcolor{ForestGreen}{EP} Gibbs, and
    \textcolor{Blue}{naive} Gibbs.
    (Left) runtime per iteration vs. number of series $N$,
    (center) NMI vs. iteration,
    (right) NMI vs. average runtime.
    The mean value and error bars (st.dev.) are over 20 trials.}
    \label{fig:synth_loglike}
    \vskip -0.2in
\end{figure*}

For synthetic data drawn from the model of Sec.~\ref{sec:tscluster}, we first demonstrate that our approximate collapsed sampler \texttt{EP Gibbs} is competitive with \texttt{naive Gibbs}' running time and with \texttt{collapsed Gibbs}' mixing rate. We simulate data using $T = 200$, $K = 20$, $\sigma_x^2 =
0.01$, $\sigma_y^2 = 1$, $a_i = 0.95$, and $\lambda = 1$. Aside from $K$, we treat all parameters as unknown in our sampling.

For our first experiment, in Fig.~\ref{fig:synth_loglike} (left) we compare the runtime per iteration as the number of series $N$, and thus number of series per cluster, varies. We clearly see that \texttt{collapsed Gibbs} scales super-linearly, while the other two methods have linear scaling. This validates that \texttt{collapsed Gibbs} is intractable for large datasets and motivates considering faster approximate samplers.

For our second experiment,
we fix $N = 300$ and measure the performance of all three samplers in terms of log-likelihood versus Gibbs iteration.
From Fig.~\ref{fig:synth_loglike} (center),
we see that on average,
\texttt{collapsed Gibbs} and our \texttt{EP Gibbs} samplers both mix quickly to a higher
log-likelihood than \texttt{naive Gibbs}, which slowly explores its high dimensional parameter space and is sensitive to local modes.
Importantly, when scaling the $x$-axis by the average runtime per iteration of 
each method, we clearly see in Fig.~\ref{fig:synth_loglike} (right) that our
\texttt{EP Gibbs} sampler handily outperforms both competitors.  \texttt{Collapsed Gibbs} is
particularly poor on these axes because of the high per-iteration runtime.
Trace plots and box plots of model parameters, rather than resulting log-likelihood, are
provided in the Appendix~\ref{app:traceplots} and show that the approximate Gibbs sampler produces similar results to Gibbs in terms of sampled mean and variance of parameters.



To demonstrate the accuracy of our approximate sampler on real time series data,
we replicate the experiment of Ren et al.~\cite{ren2015achieving} to
predict house prices in the city of Seattle.
The data consists of 124,480 housing transactions in 140 census tracts
(series) of Seattle from 1997 to 2013,
partitioned into a 75-25 train test split stratified by series.
Each transaction consists of a sales price, our prediction target, and
house-specific covariates such as `lot square-feet' or `number of bathrooms'.
We first remove a global trend and jointly fit
the time series clustering model with
series-specific regressions on individual transaction covariates.
Full details can be found in the Appendix~\ref{app:housing_data}.

We compare fitting this model using our approximate sampler to the collapsed
Gibbs sampler of Ren et al. using the same error metrics as in that paper:
root mean-squared error (RMSE) in price, and mean / median / 90th percentile
of absolute percent error (APE).

\begin{figure}[htb]
   \vskip 0.05in
    \centering
    \begin{minipage}[t]{.45\textwidth}
    \centering
        \includegraphics[width=\textwidth]{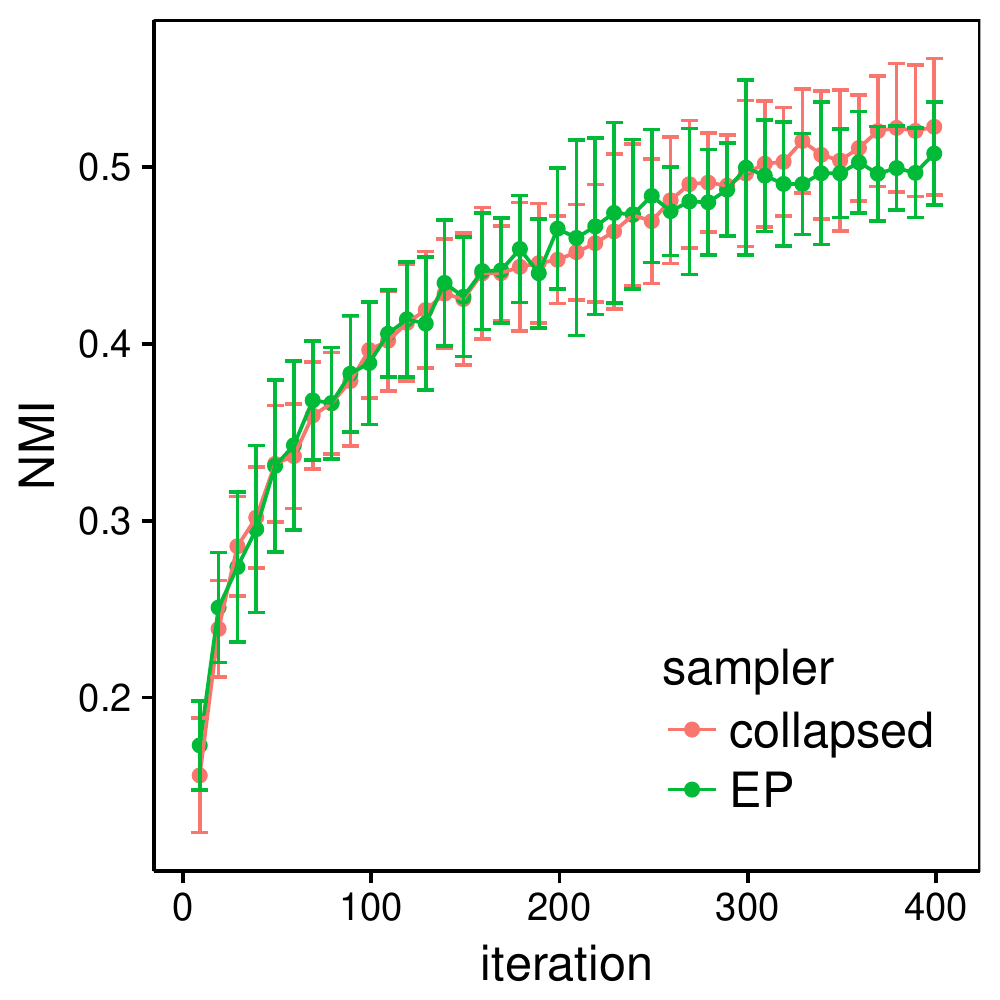}
    \end{minipage}
    \hspace{0.05in}
    \begin{minipage}[t]{.45\textwidth}
    \centering
        \includegraphics[width=\textwidth]{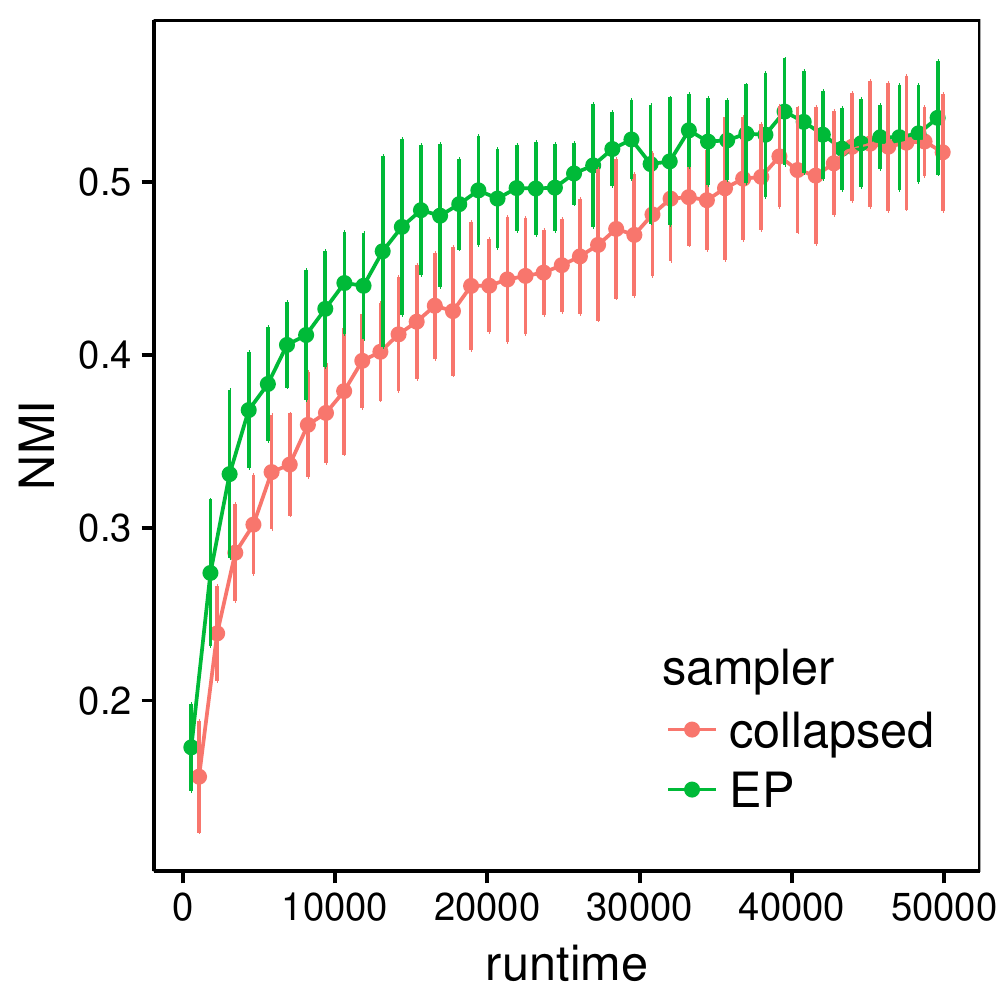}
    \end{minipage}
\vspace{-0.15in}
    \caption{\small Comparisons to 
    the collapsed sampler MAP assignment on the housing data:
    \textcolor{BrickRed}{collapsed} Gibbs and
    our \textcolor{ForestGreen}{EP} Gibbs.
    (Left) NMI vs iteration, (right) NMI vs runtime.
    }
    \label{fig:zillow_compare}
    \vskip -0.2in
\end{figure}

The performance of our approximate sampler \texttt{EP Gibbs} and the \texttt{collapsed Gibbs} sampler are 
presented in Table~\ref{tab:housing}; 
we include NMI comparisons to the MAP of the \texttt{collapsed Gibbs} in Fig.~\ref{fig:zillow_compare}. 
We see that both algorithms for time series
clustering produce similar results on all metrics (within a standard
deviation). However, \texttt{EP Gibbs} achieves superior performance much more
rapidly. As such, we view our algorithm as an attractive alternative in this
case. Furthermore, note that our gains would only increase with the size of the
dataset, e.g., number of regions $N$, a limitation of~\cite{ren2015achieving}.

\vspace{-0.1pt}
\begin{table}[h]
    \caption{Test metrics on housing data averaged over 10 trials.
    Parenthetical values are one standard deviation.}
    \label{tab:housing}
    \begin{center}
    \begin{tabular}{lcc}
      \hline
      metric & collapsed & EP \\
      \hline
      RMSE & 125280 (50) & 125280 (80) \\ 
      Mean APE & 16.20 (0.01) & 16.20 (0.01) \\ 
      Median APE & 12.07 (0.01) & 12.07 (0.01) \\ 
      90th APE & 34.17 (0.07) & 34.22 (0.05) \\
      Runtime & 121.6 (8.1) & 62.8 (3.7)\\
      \hline
    \end{tabular}
    \end{center}
\end{table}

%

\presec
\section{Conclusion}
\postsec
We presented a framework for constructing approximate collapsed Gibbs samplers for efficient inference in complex clustering models. The key idea is to approximately marginalize the nuisance variables by using EP to approximate the conditional distributions of the variables with an individual observation removed; by approximating this conditional, the required integral becomes tractable in a much wider range of scenarios than that of conjugate models.  Our use of this EP approximation takes two steps from its traditional use: (1) we approximate a (nearly) full conditional rather than directly targeting the posterior, and (2) our targeted conditional changes as we sample the cluster assignment variables.  For the latter, we provided a brief analysis and demonstrated the impact of the changing target, drawing parallels to previously proposed samplers that use stale sufficient statistics. 

We demonstrated how to apply our EP-based approximate sampling approach in two applications:
mixtures of Student-$t$ distributions and time series clustering.
Our experiments demonstrate that our EP approximate collapsed samplers mix more rapidly
than naive Gibbs, while being computationally scalable and analytically tractable.
We expect this method to provide the greatest benefit when approximately collapsing large parameter spaces.

There are many interesting directions for future work,
including deriving bounds on the asymptotic convergence of our approximate
sampler~\cite{pillai2014ergodicity, dinh2017convergence},
considering different likelihood approximation update rules such as \emph{power EP}~\cite{minka2004power},
and extending our idea of approximately integrating out variables to other samplers.
For the analysis, \cite{dehaene2015expectation} showed that EP with Gaussian approximations is exact in the large data limit; one could extend these results to consider the case of data being allocated amongst \emph{multiple} clusters.  
Another interesting direction is to explore our EP-based approximate collapsing within the context of variational inference, 
possibly extending the set of models for which collapsed variational Bayes~\cite{teh2007collapsed} is possible. 
Finally, there are many ways in which our algorithm could be made even more scalable through distributed, asynchronous implementations, such as in~\cite{ahmed2012scalable}.

%


\section*{Acknowledgements} 
We would like to thank Nick Foti, You “Shirley” Ren and
Alex Tank for helpful discussions.
This paper is based upon work supported by the NSF
CAREER Award IIS-1350133

This paper is an extension of our previous workshop paper~\cite{aicher2016scalable}.


\bibliography{bib}

\begin{thebibliography}{40}
\providecommand{\natexlab}[1]{#1}
\providecommand{\url}[1]{\texttt{#1}}
\expandafter\ifx\csname urlstyle\endcsname\relax
  \providecommand{\doi}[1]{doi: #1}\else
  \providecommand{\doi}{doi: \begingroup \urlstyle{rm}\Url}\fi

\bibitem[Ahmed et~al.()Ahmed, Aly, Gonzalez, Narayanamurthy, and
  Smola]{ahmed2012scalable}
A.~Ahmed, M.~Aly, J.~Gonzalez, S.~Narayanamurthy, and A.~Smola.
\newblock Scalable inference in latent variable models.
\newblock In \emph{International conference on Web search and data mining
  (WSDM)}, volume~51, pages 1257--1264.

\bibitem[Aicher and Fox(2016)]{aicher2016scalable}
C.~Aicher and E.~B. Fox.
\newblock Scalable clustering of correlated time series using expectation
  propagation.
\newblock \emph{2nd SIGKDD Workshop on Mining and Learning from Time Series},
  2016.

\bibitem[Airoldi et~al.(2008)Airoldi, Blei, Fienberg, and
  Xing]{airoldi2008mixed}
E.~M. Airoldi, D.~M. Blei, S.~E. Fienberg, and E.~P. Xing.
\newblock Mixed membership stochastic blockmodels.
\newblock \emph{Journal of Machine Learning Research}, 9\penalty0
  (Sep):\penalty0 1981--2014, 2008.

\bibitem[Andrews and Mallows(1974)]{andrews1974scale}
D.~F. Andrews and C.~L. Mallows.
\newblock Scale mixtures of normal distributions.
\newblock \emph{Journal of the Royal Statistical Society. Series B
  (Methodological)}, pages 99--102, 1974.

\bibitem[Barber et~al.(2011)Barber, Cemgil, and Chiappa]{barber2011inference}
D.~Barber, A.~T. Cemgil, and S.~Chiappa.
\newblock Inference and estimation in probabilistic time series models.
\newblock \emph{Bayesian Time Series Models}, 2011.

\bibitem[Bishop(2004)]{bishop2004robust}
C.~M. Bishop.
\newblock Robust bayesian mixture modelling.
\newblock Citeseer, 2004.

\bibitem[Bishop(2006)]{bishop2006pattern}
C.~M. Bishop.
\newblock Pattern recognition.
\newblock \emph{Machine Learning}, 2006.

\bibitem[Blei and Lafferty(2007)]{blei2007correlated}
D.~M. Blei and J.~D. Lafferty.
\newblock A correlated topic model of science.
\newblock \emph{The Annals of Applied Statistics}, pages 17--35, 2007.

\bibitem[Blei et~al.(2003)Blei, Ng, and Jordan]{blei2003latent}
D.~M. Blei, A.~Y. Ng, and M.~I. Jordan.
\newblock Latent dirichlet allocation.
\newblock \emph{Journal of machine Learning research}, 3\penalty0
  (Jan):\penalty0 993--1022, 2003.

\bibitem[Blei et~al.(2017)Blei, Kucukelbir, and McAuliffe]{blei2017variational}
D.~M. Blei, A.~Kucukelbir, and J.~D. McAuliffe.
\newblock Variational inference: A review for statisticians.
\newblock \emph{Journal of the American Statistical Association}, \penalty0
  (just-accepted), 2017.

\bibitem[Damlen et~al.(1999)Damlen, Wakefield, and Walker]{damlen1999gibbs}
P.~Damlen, J.~Wakefield, and S.~Walker.
\newblock Gibbs sampling for bayesian non-conjugate and hierarchical models by
  using auxiliary variables.
\newblock \emph{Journal of the Royal Statistical Society: Series B (Statistical
  Methodology)}, 61\penalty0 (2):\penalty0 331--344, 1999.

\bibitem[Dehaene and Barthelm{\'e}(2015)]{dehaene2015expectation}
G.~Dehaene and S.~Barthelm{\'e}.
\newblock Expectation propagation in the large-data limit.
\newblock \emph{arXiv preprint arXiv:1503.08060}, 2015.

\bibitem[Dinh et~al.(2017)Dinh, Rundell, and Buzzard]{dinh2017convergence}
V.~Dinh, A.~E. Rundell, and G.~T. Buzzard.
\newblock Convergence of griddy gibbs sampling and other perturbed markov
  chains.
\newblock \emph{Journal of Statistical Computation and Simulation}, 87\penalty0
  (7):\penalty0 1379--1400, 2017.

\bibitem[Dunson(2000)]{dunson2000bayesian}
D.~B. Dunson.
\newblock Bayesian latent variable models for clustered mixed outcomes.
\newblock \emph{Journal of the Royal Statistical Society: Series B (Statistical
  Methodology)}, 62\penalty0 (2):\penalty0 355--366, 2000.

\bibitem[Escobar and West(1995)]{escobar1995bayesian}
M.~D. Escobar and M.~West.
\newblock Bayesian density estimation and inference using mixtures.
\newblock \emph{Journal of the american statistical association}, 90\penalty0
  (430):\penalty0 577--588, 1995.

\bibitem[Fan and Bouguila(2014)]{fan2014non}
W.~Fan and N.~Bouguila.
\newblock Non-gaussian data clustering via expectation propagation learning of
  finite dirichlet mixture models and applications.
\newblock \emph{Neural processing letters}, 39\penalty0 (2):\penalty0 115--135,
  2014.

\bibitem[Heskes and Zoeter(2002)]{heskes2002expectation}
T.~Heskes and O.~Zoeter.
\newblock Expectation propagation for approximate inference in dynamic bayesian
  networks.
\newblock In \emph{Proceedings of the Eighteenth conference on Uncertainty in
  artificial intelligence}, pages 216--223. Morgan Kaufmann Publishers Inc.,
  2002.

\bibitem[Hoffman et~al.(2013)Hoffman, Blei, Wang, and
  Paisley]{hoffman2013stochastic}
M.~D. Hoffman, D.~M. Blei, C.~Wang, and J.~Paisley.
\newblock Stochastic variational inference.
\newblock \emph{The Journal of Machine Learning Research}, 14\penalty0 (1),
  2013.

\bibitem[Johndrow et~al.(2015)Johndrow, Mattingly, Mukherjee, and
  Dunson]{johndrow2015approximations}
J.~E. Johndrow, J.~C. Mattingly, S.~Mukherjee, and D.~Dunson.
\newblock Approximations of markov chains and bayesian inference.
\newblock \emph{arXiv preprint arXiv:1508.03387}, 2015.

\bibitem[Kingma and Welling(2013)]{kingma2013auto}
D.~P. Kingma and M.~Welling.
\newblock Auto-encoding variational bayes.
\newblock \emph{arXiv preprint arXiv:1312.6114}, 2013.

\bibitem[LeCun and Cortes(2010)]{lecunmnisthandwrittendigit2010}
Y.~LeCun and C.~Cortes.
\newblock {MNIST} handwritten digit database.
\newblock 2010.
\newblock URL \url{http://yann.lecun.com/exdb/mnist/}.

\bibitem[Li et~al.()Li, Andersen, and Park]{li2014scaling}
M.~Li, D.~G. Andersen, and J.~W. Park.
\newblock Scaling distributed machine learning with the parameter server.

\bibitem[Liu(1996)]{liu1996bayesian}
C.~Liu.
\newblock Bayesian robust multivariate linear regression with incomplete data.
\newblock \emph{Journal of the American Statistical Association}, 91\penalty0
  (435):\penalty0 1219--1227, 1996.

\bibitem[Liu(2008)]{liu2008monte}
J.~S. Liu.
\newblock \emph{Monte Carlo strategies in scientific computing}.
\newblock Springer Science \& Business Media, 2008.

\bibitem[Ma et~al.(2015)Ma, Chen, and Fox]{ma2015complete}
Y.-A. Ma, T.~Chen, and E.~Fox.
\newblock A complete recipe for stochastic gradient mcmc.
\newblock In \emph{Advances in Neural Information Processing Systems}, pages
  2917--2925, 2015.

\bibitem[Minka(2004)]{minka2004power}
T.~Minka.
\newblock Power ep.
\newblock Technical report, Technical report, Microsoft Research, Cambridge,
  2004.

\bibitem[Minka(2001)]{minka2001expectation}
T.~P. Minka.
\newblock Expectation propagation for approximate bayesian inference.
\newblock In \emph{Proceedings of the Seventeenth conference on Uncertainty in
  artificial intelligence}. Morgan Kaufmann Publishers Inc., 2001.

\bibitem[Murray and Ghahramani(2004)]{murray2004bayesian}
I.~Murray and Z.~Ghahramani.
\newblock Bayesian learning in undirected graphical models: approximate mcmc
  algorithms.
\newblock In \emph{Proceedings of the 20th conference on Uncertainty in
  artificial intelligence}, pages 392--399. AUAI Press, 2004.

\bibitem[Neal et~al.(2011)]{neal2011mcmc}
R.~M. Neal et~al.
\newblock Mcmc using hamiltonian dynamics.
\newblock \emph{Handbook of Markov Chain Monte Carlo}, 2:\penalty0 113--162,
  2011.

\bibitem[Peel and McLachlan(2000)]{peel2000robust}
D.~Peel and G.~J. McLachlan.
\newblock Robust mixture modelling using the t distribution.
\newblock \emph{Statistics and computing}, 10\penalty0 (4):\penalty0 339--348,
  2000.

\bibitem[Pillai and Smith(2014)]{pillai2014ergodicity}
N.~S. Pillai and A.~Smith.
\newblock Ergodicity of approximate mcmc chains with applications to large data
  sets.
\newblock \emph{arXiv preprint arXiv:1405.0182}, 2014.

\bibitem[Portilla et~al.(2003)Portilla, Strela, Wainwright, and
  Simoncelli]{portilla2003image}
J.~Portilla, V.~Strela, M.~J. Wainwright, and E.~P. Simoncelli.
\newblock Image denoising using scale mixtures of gaussians in the wavelet
  domain.
\newblock \emph{IEEE Transactions on Image processing}, 12\penalty0
  (11):\penalty0 1338--1351, 2003.

\bibitem[Ren et~al.(2015)Ren, Fox, and Bruce]{ren2015achieving}
Y.~Ren, E.~B. Fox, and A.~Bruce.
\newblock Achieving a hyperlocal housing price index: Overcoming data sparsity
  by bayesian dynamical modeling of multiple data streams.
\newblock \emph{arXiv preprint arXiv:1505.01164}, 2015.

\bibitem[Ritter and Tanner(1992)]{ritter1992facilitating}
C.~Ritter and M.~A. Tanner.
\newblock Facilitating the gibbs sampler: the gibbs stopper and the
  griddy-gibbs sampler.
\newblock \emph{Journal of the American Statistical Association}, 87\penalty0
  (419):\penalty0 861--868, 1992.

\bibitem[Seeger(2005)]{seeger2005expectation}
M.~Seeger.
\newblock Expectation propagation for exponential families.
\newblock Technical report, 2005.

\bibitem[Snijders and Nowicki(1997)]{snijders1997estimation}
T.~A. Snijders and K.~Nowicki.
\newblock Estimation and prediction for stochastic blockmodels for graphs with
  latent block structure.
\newblock \emph{Journal of classification}, 14\penalty0 (1):\penalty0 75--100,
  1997.

\bibitem[Teh et~al.(2007)Teh, Newman, and Welling]{teh2007collapsed}
Y.~W. Teh, D.~Newman, and M.~Welling.
\newblock A collapsed variational bayesian inference algorithm for latent
  dirichlet allocation.
\newblock In \emph{Advances in neural information processing systems}, pages
  1353--1360, 2007.

\bibitem[Teh et~al.(2015)Teh, Hasenclever, Lienart, Vollmer, Webb,
  Lakshminarayanan, and Blundell]{teh2015distributed}
Y.~W. Teh, L.~Hasenclever, T.~Lienart, S.~Vollmer, S.~Webb,
  B.~Lakshminarayanan, and C.~Blundell.
\newblock Distributed bayesian learning with stochastic natural-gradient
  expectation propagation and the posterior server.
\newblock \emph{arXiv preprint arXiv:1512.09327}, 2015.

\bibitem[Van~Dyk and Park(2008)]{van2008partially}
D.~A. Van~Dyk and T.~Park.
\newblock Partially collapsed gibbs samplers: Theory and methods.
\newblock \emph{Journal of the American Statistical Association}, 103\penalty0
  (482), 2008.

\bibitem[Vinh et~al.(2010)Vinh, Epps, and Bailey]{vinh2010information}
N.~X. Vinh, J.~Epps, and J.~Bailey.
\newblock Information theoretic measures for clusterings comparison: Variants,
  properties, normalization and correction for chance.
\newblock \emph{The Journal of Machine Learning Research}, 11, 2010.

\end{thebibliography}
\bibliographystyle{abbrvnat}

\clearpage
\appendix
\renewcommand{\theequation}{\Alph{section}.\arabic{equation}}
\begin{center}
\LARGE \textbf{Appendix}
\end{center}

\section{Mixture of Multivariate Student-$t$}
\label{app:student}
This section provides additional details for
Sec.~\ref{sec:student} on multivariate Student-$t$ distributions (MVT).
We first provide the details for the naive and blocked (partially collapsed)
Gibbs sampler based on data augmentation.
We then provide the details on how to approximate the collapsed log-likelihood
and moments required for our EP approximation.

\subsection{Naive Sampler Steps}
For notation, we will let $2\alpha$ be the degrees of freedom of the MVT distribution and reserve $\nu$ for the degrees of freedom in the inverse-Wishart distribution.

\subsubsection*{Sampling $z$}
\begin{equation}
    \label{eq:naive_z_student_step}
    \Pr( z_i \, | \, z_{-i}, \mu, \Sigma, u, y) \propto \mathcal{N}(y_i \, | \, \mu_{z_i}, \Sigma_{z_i}/u_{i, z_i}) \Pr(z_i \, | \, z_{-i})
\end{equation}
which can be evaluated for each $z_i = k$ and then normalized.

\subsubsection*{Sampling $\mu, \Sigma$}
\begin{equation}
\Pr(\mu_k, \Sigma_k \, | \, z, u, y) 
    = \mathcal{NIW}(\mu_k, \Sigma_k \, | \, \mu_p, \kappa_p, \nu_p, \Psi_p)
    = \mathcal{N}(\mu_k \, | \, \mu_p , \Sigma_k/\kappa_p) \cdot \mathcal{IW}(\Sigma_k \, | \, \nu_p, \Psi_p)
\label{eq:mu_Sigma_post}
\end{equation}
where $\mathcal{IW}$ is the inverse-Wishart distribution
\begin{equation*}
\mathcal{IW}(\Sigma \, | \, \nu, \Psi) = \frac{|\Psi|^{\nu/2}}{2^{\nu d/2} \Gamma_d(\nu/2)} |\Sigma|^{-\frac{\nu + d + 1}{2}} e^{-\frac{1}{2}\tr(\Psi X^{-1})}
\end{equation*}
and
\begin{align*}
\mu_p &= (\kappa_0\mu_0 + \sum_{z_i = k} u_i y_i)/\kappa_p \\
\kappa_p &= \kappa_0 + \sum_{z_i = k} u_i \\
\nu_p &= \nu_0 + \sum_{z_i =k} 1 \\
\Psi_p &= \Psi_0 + \kappa_0 \mu_0 \mu_0^T + \sum_{z_i = k} u_i y_i y_i^T - \kappa_p \mu_p \mu_p^T
\end{align*}
when $(\mu_0, \kappa_0, \nu_0, \Psi_0)$ are the parameters of the prior.

\subsubsection*{Sampling $u$}
\begin{equation}
     \Pr(u_{i,k} \, | \, z, \mu, \Sigma, y) = \begin{cases}
     \Gamma(u_{i,k} \, | \, \alpha, \alpha) & \text{ if } z_i \neq k \\
     \Gamma(u_{i,k} \, | \, \alpha_1^*, \alpha_2^*)& \text{ if } z_i = k 
     \end{cases}
\end{equation}
where $\alpha_1^* = \alpha + d/2$ and $\alpha_2^* = \alpha  + (y_i - \mu_k)^T \Sigma_k^{-1}(y_i - \mu_k)$.

For the correctness of the sampler, 
we must to sample a separate $u_{i,k}$ for each observation-cluster pair ($i,k$).

\subsection{Blocked Sampler Steps}
Given a conjugate prior for $\mu, \Sigma$ (normal inverse-Wishart),
the posterior over $\mu, \Sigma$ for fixed $z$ and $u$ is normal inverse-Wishart (see Eq.~\eqref{eq:mu_Sigma_post}).

Therefore we can integrate out $\mu$ and $\Sigma$ in the likelihood of Eq.~\eqref{eq:naive_z_student_step} to obtain
\begin{equation*}
    \Pr( z_i \, | \, z_{-i}, u, y) \propto \int
    \mathcal{N}\left(y_i \, | \, \mu_{z_i}, \dfrac{\Sigma_{z_i}}{u_{i, z_i}}\right) q(\mu_{z_i}, \Sigma_{z_i} \, | \, y_{-i})
 \times \Pr(z_i \, | \, z_{-i}) \, d\mu_{z_i} d\Sigma_{z_i}
\end{equation*}
where $q(\mu_{z_i}, \Sigma_{z_i} \, | \, y_{-i}) = \mathcal{NIW}(\mu_{z_i}, \Sigma_{z_i} \, | \, \mu_p, \kappa_p, \nu_p, \Psi_p)$ is the NIW posterior calculated without observation $i$.

Taking the integral, we obtain
\begin{equation}
\Pr( z_i | z_{-i}, u, y) = t_{2\alpha_p}\left(y_i \, | \, \mu_p, \Sigma_p\right)
\label{eq:mvt_constant}
\end{equation}
where $t$ is a MVT distribution with mean $\mu_p$, covariance matrix $\Sigma_p =\frac{(\kappa_p + u_{i,z_i}) \Psi_p}{(\kappa_p \cdot u_{i, z_i})(\nu_p - d + 1)} $ and degrees of freedom $2\alpha_p = \nu_p - d + 1$.

\subsection{EP Approximate Log-likelihood}
We now present how to approximate the collapsed likelihood,
approximating $p(\mu, \Sigma | y, z, u)$ with a normal inverse-Wishart $q(\mu,\Sigma)$.

The normalizing constant (a.k.a. the likelihood approximation) for fixed $u$ is given by the block sampler where our prior is our cavity distribution $q(\mu_{z_i},  \Sigma_{z_i} \, | \, y_{-i})$.

Therefore we can (tractably) estimate the normalizing constant by numerically integrating out (the univariate) $u_{i,k}$: the integrand is a MVT evaluated at $y_i$ with changing variance $\Sigma_p(u_{i,k})$ (see~\eqref{eq:mvt_constant}).

\subsection{EP Moment Update}
To update our EP approximation $q(\mu, \Sigma)$ we must calculate the moments of the sufficient statistics of $\mu, \Sigma$.
For a normal inverse-Wishart the sufficient statistics and their moments are
\begin{equation*}
    \begin{aligned}
    T_1 &= \Sigma^{-1}\mu \\
    T_2 &= \mu^T\Sigma^{-1}\mu \\
    T_3 &= \nu \Sigma^{-1}\\
    T_4 &= -\log | \Sigma |
    \end{aligned}
    \enspace
    \Rightarrow
    \enspace
    \begin{aligned}
    \E[T_1] &= \nu \Psi^{-1} \mu \\
    \E[T_2] &= \nu \mu^T\Psi^{-1} \mu + d/\kappa \\
    \E[T_3] &= \nu \Psi^{-1} \\
    \E[T_4] &= \psi_d(\nu/2) + d\log 2 - \log | \Psi|
    \end{aligned}
\end{equation*}
where $\psi_d$ is the multivariate digamma function.

If $u$ was a point mass, then the titled moments would be straightforward to calculate; just plug in the appropriate $\mu, \kappa, \nu, \Psi$ as function of $u$.
Because we must integrate with respect to $\Gamma(u \, | \, \alpha, \alpha)$, we can approximate the integral with a Riemann sum.
The moments can be calculated efficiently for a vector of $u$ by recognizing they all differ by at most a rank-one update to the parameters $\mu, \kappa, \nu, \Psi$ and using the Woodbury matrix identity and determinant matrix lemma.

All that remains is to solve for the new posterior parameters by matching moments.
This can be done by solving a system of equations.
Note that for $\nu$,
we must solve a 1-dimensional root finding problem to handle the digamma function $\psi_d$, which can be done quickly.

\section{Time Series Clustering}
\label{app:tscluster}

This section provides additional details for
Sec.~\ref{sec:tscluster} on time series clustering.
We describe how to calculate log-likelihoods using the Kalman smoother
and how to calculate the posterior moments of $\eta$ for our EP approximation.

We consider time series clustering model is given by
Eq.~\eqref{eq:eta_model}.
For the rest of this section, we assume conditioning on all parameters except
$x$, $z$, and $\eta$ (i.e. $a$, $\lambda$, $\sigma_x^2$, $\sigma_y^2$),
unless otherwise noted.
The Gibbs sampling distribution for these other likelihood parameters can be 
found in the appendix of Ren et al.~\cite{ren2015achieving}.

\subsection{Naive Log-likelihood}
Collapsing only $x$,
the naive Gibbs sampler likelihood for $z_i$ is given by
Eq.~\eqref{eq:ts_naive_z_like}, which is
\begin{equation}
    \label{supp-eq:ts_naive_z_like}
p_\phi(y_i | z_i, \eta) =
    \int \prod_{t = 1}^T \Pr(y_{i,t} |  x_{i,t}) \Pr(x_{i,t} |  x_{i,t-1},
    \eta_{z_i,t}) \, d x_t
\end{equation}
By assumption, both 
the conditional distribution of $y_{i,t}$ given $x_{i,t}$ 
and the conditional distribution of $x_{i,t}$ given $x_{i,t-1}$ are 
Gaussian
\begin{align}
    \label{supp-eq:ts_naive_1}
    \Pr(y_{i,t} \, | \, x_{i,t}) &= 
        \mathcal{N}(y_{i,t} \, | \, x_{i,t}, \sigma_y^2) \\
    \label{supp-eq:ts_naive_2}
    \Pr(x_{i,t} \, | \, x_{i,t-1}) &=
        \mathcal{N}(x_{i,t} \, | \, a_i x_{i,t} + \lambda_i\eta_{z_i,t},
        \sigma_x^2)
    \enspace.
\end{align}

The likelihood Eq.~\ref{supp-eq:ts_naive_z_like} is then calculated using the 
Kalman filter~\cite{bishop2006pattern},
which consists of iteratively applying `predict' and `update' steps.
Due to the perturbations $\lambda_i \eta_{z_i, t}$ there is a slight adjustment
in the predict step~\cite{ren2015achieving}.

Let $\mu_{t|t-1}, \sigma^2_{t|t-1}$ denote the predictive mean and variance of
$x_{i,t}$ given $y_{i,1:t-1}, \eta_{z_i}$ and
let $\mu_{t | t}, \sigma^2_{t | t}$ denote the filtered mean and variance of
$x_{i,t}$ given $y_{i,1:t}, \eta_{z_i}$.
We can iteratively calculate the predictive and filtered parameters by applying
`predict' and `update' steps.

The predict step is 
\begin{align}
    \nonumber
    \mu_{t | t-1} &= a_i \mu_{t-1|t-1} + \lambda_i \eta_{z_i, t} \\
    \sigma^2_{t|t-1} &= a_i^2 \sigma^2_{t-1|t-1} + \sigma_x^2 \enspace.
    \label{supp-eq:predict_step_univariate}
\end{align}  

The update step is
\begin{align}
    \nonumber
    \mu_{t | t} &= \mu_{t|t-1} + K_t \cdot (y_{i,t} - \mu_{t|t-1}) \\
    \sigma^2_{t|t-1} &= (1 - K_t) \sigma^2_{t|t-1} \enspace.
    \label{supp-eq:update_step_univariate}
\end{align}

where $K_t$ is \textit{Kalman} gain
\begin{equation*}
    K = \sigma^2_{t|t-1}/(\sigma^2_{t|t-1} + \sigma^2_{y_i} )\enspace.
\end{equation*}

We calculate the log-likelihood of $y_i$ by factorizing over time
\begin{equation}
    \log \Pr(y_i \, | \, z_i , \eta) = \sum_{t = 1}^T \log \Pr(y_{i,t} \,|\,
    y_{i,<t}, z_i, \eta) \enspace,
\end{equation}
where $\Pr(y_{i,t} | y_{i, <t}, z_i, \eta)$ is the Gaussian
\begin{equation*}
    \Pr(y_{i,t} | y_{i, <t}, z_i, \eta)
    = \int \Pr(y_{i,t} | x_{i,t}) \Pr(x_{i,t} | y_{i,<t}, z_i, \eta) \,dx_{i,t}
    = \mathcal{N}(y_{i,t} | \mu_{t|t-1}, \sigma^2_{t|t-1} +
    \sigma^2_{y_i}) \enspace.
\end{equation*}

\subsection{Collapsed Log-likelihood}
Collapsing both $x$ and $\eta$,
the collapsed Gibbs sampler likelihood for $z_i$ is
\begin{equation}
    \Pr(y_i | z, y_{-i}) = 
    \int \! \int  \prod_{t = 1}^T \Pr(y_{i,t} |  x_{i,t}) \Pr(x_{i,t} |  x_{i,t-1},
    \eta_{z_i,t})
    \times \Pr(\eta_{z_i} \, | \, y_{-i}, z_{-i}) \ d \eta \ d x_{t:T} \enspace.
    \label{supp-eq:ts_collapsed_int}
\end{equation}
Although the distribution $\Pr(\eta_{z_i} | y_{-i}, z_{-i})$ is known to be 
a $T$-dimensional multivariate Gaussian,
computing its parameters and directly evaluating this integral,
Eq.~\eqref{supp-eq:ts_collapsed_int},
is computationally prohibitive even for moderate sizes of $T$:
inverting the covariance matrix requires $O(T^3)$ computation.

Ren et al.~\cite{ren2015achieving}
exploited the time-series structure of Fig.~\ref{fig:GM} (bottom-right)
to calculate the collapsed likelihood by factorizing over time
\begin{equation}
    \label{supp-eq:ts_collapsed_z_like}
    \Pr(y_i \, | \, z, y_{-i}) = 
    \prod_{t = 1}^T \Pr(y_{i,t} \, | \, y_{-i,t}, y_{1:t-1}, z) \enspace,
\end{equation}
where each conditional distribution in the product 
$\Pr(y_{i,t} | y_{-i,t}, y_{1:t-1})$
can be calculated from the
joint distribution
\begin{equation*}
    \Pr(y_t \ | \ y_{1:t-1}) = \int \Pr(y_t \, | \, x_t) \Pr(x_t \, | \,
    y_{1:t-1}) \ dx_t \enspace.
\end{equation*}
Here, we let $y_t$ and $x_t$ denote the vector of values at time $t$ for series
in cluster $z_i$.
Recall that the values of other series $z_j \neq z_i$ are conditionally
independent.
The predictive distribution $\Pr(x_t \, | \, y_{1:t-1})$ is calculated by 
the predict step of the multivariate generalization of the
Kalman filter~\cite{bishop2006pattern, ren2015achieving}.

Let $\mu_{t|t-1}, \Sigma_{t|t-1}$ denote the predictive mean and variance of
$x_{t}$ given $ y_{1:t-1}$ and
let $\mu_{t | t}, \Sigma_{t | t}$ denote the filtered mean and variance of
$x_{t}$ given $y_{1:t}$.
We can iteratively calculate the predictive and filtered parameters by applying
`predict' and `update' steps.

The predict step is 
\begin{align}
    \nonumber
    \mu_{t | t-1} &= A \mu_{t-1|t-1} \\
    \Sigma_{t|t-1} &= A \Sigma_{t-1|t-1} A^T + \mathbb{I}\sigma_x^2 + 
    \lambda \lambda^T \enspace,
    \label{supp-eq:predict_step_multivariate}
\end{align}
where $A = \diag(a)$.
Note that the additional covariance term $\lambda \lambda^T$ couples the series
together and is due to collapsing out $\eta_t$.

The update step is
\begin{align}
    \nonumber
    \mu_{t | t} &= \mu_{t|t-1} + K_t \cdot (y_{i,t} - \mu_{t|t-1}) \\
    \Sigma_{t|t-1} &= (\mathbb{I} - K_t) \Sigma_{t|t-1} \enspace.
    \label{supp-eq:update_step_multivariate}
\end{align}

where $K_t$ is \textit{Kalman} gain
\begin{equation*}
    K = \Sigma_{t|t-1} \cdot (\Sigma_{t|t-1} + \diag(\sigma^2_{y}) )^{-1}\enspace.
\end{equation*}

%

Note that we must solve linear systems in the update step and in calculating
the conditional likelihood.
As these linear systems are of dimension $O(N_k)$,
practical numerical solvers have a runtime complexity $O(N_k^3)$.
As a result the full runtime complexity of evaluating
Eq.~\eqref{supp-eq:ts_collapsed_z_like} is $O(T N_k^3)$ for each cluster assignment
$z_i = k$.

\subsection{EP Approximate Log-likelihood}
To approximate the collapsed likelihood,
we use EP to keep track of a 
diagonal Gaussian approximations $q(\eta_{k} | z)$ for
$\Pr(\eta_{k} | y, z)$.
Because $q$ is diagonal, it factorizes over time
\begin{equation}
    q(\eta_{k} | z) = \prod_{t = 1}^T
    \mathcal{N}(\eta_{k, t} \, | \, \mu_{k,t}, \sigma_{k,t}^2)
    \enspace.
\end{equation}

To calculate the cavity distribution $q(\eta_{k} | z_{-i})$,
we remove the site approximation $\tilde{f}_i(\eta_k)$ from $q(\eta_k | z)$.
This can be done by subtracting the natural parameters (mean-precision and
precision).

If $\tilde{f}_i(\eta_k) = C_i \mathcal{N}(\eta_k | \mu_i, \sigma_i^2)$,
then the mean and diagonal variance of the cavity distribution is
\begin{align}
    \label{supp-eq:cavity_ts_mean}
    \mu^{(-i)}_{k} &= {\sigma^2}_{k}^{(-i)} \cdot
    \left( \mu_k / {\sigma^2}_{k} - \mu_i / {\sigma^2}_i \right) \\
    \label{supp-eq:cavity_ts_var}
    {\sigma^2}_{k}^{(-i)} &= 
    \left(1/{\sigma^2}_{k} - 1/{\sigma^2}_i\right)^{-1}
\enspace.
\end{align}

Our approximation for the collapsed likelihood is
\begin{align}
    \nonumber
    \Pr(y_i | z, y_{-i})
    &\approx \int \prod_{t = 1}^T  \Big[
    \int \Pr(y_{i,t} | x_{i,t}) \Pr(x_{i,t} |  x_{i,t-1}, \eta_{z_i,t}) \, d x_{i,t}
    \Big] \times q(\eta_{z_i} | z_{-i})
    \, d \eta
    \\
    \nonumber
    &= \int  \prod_{t = 1}^T \Big[
    \int \Pr(x_{i,t} |  x_{i,t-1}, \eta_{z_i,t}) q(\eta_{z_i, t} | z_{-i}) \, d\eta_{z_i, t}
    \Big]
    \times \Pr(y_{i,t} \, | \, x_{i,t}) \, d x_{i,t}
    \\
    \label{supp-eq:ts_ep_z_like}
    &= \int \prod_{t = 1}^T \,
        \mathcal{N}\left(x_{i,t} \, | \,
        a_i x_{i,t} + \lambda_i \mu^{(-i)}_{z_i,t} \, , \,
        \sigma_x^2 + \lambda_i^2 {\sigma^2}_{z_i,t}^{(-i)} \right)
        \times \mathcal{N}(y_{i,t} \, | \, x_{i,t}, \sigma_y^2)  \, d x_{i,t}
    \enspace.
\end{align}
Note that the integral product of Eq.~\eqref{supp-eq:ts_ep_z_like} (second line) 
is similar in form to the naive likelihood (Eq.~\eqref{supp-eq:ts_naive_z_like});
both take the form
\begin{equation*}
\int \prod_{t=1}^T \Pr(y_{i,t} | x_{i,t}) \Pr(x_{i,t} | x_{i,t-1}) \, dx_{i,t} \enspace.
\end{equation*}
The only difference (between Eq.~\eqref{supp-eq:ts_ep_z_like} (third line) and Eq.~\eqref{supp-eq:ts_naive_2})  is that latent process $x_i$ is `smoothed' by marginalizing over the cavity distribution of $\eta_{z_i}$, the variance is a bit larger
($\lambda_i^2 {\sigma^2}_{z_i, t}^{(-i)}$) and
the mean shift uses ${\mu}_{z_i, t}^{(-i)}$, instead of using
the point estimate $\eta_{z_i}$ from the previous iteration.
Therefore, we can calculate our approximation with the univariate Kalman filter
(Eqs.\eqref{supp-eq:predict_step_univariate} and \eqref{supp-eq:update_step_univariate})
in $O(T)$ time.

Our modified predict step (replacing Eq.~\eqref{supp-eq:predict_step_univariate}) is
\begin{align}
    \nonumber
    \mu_{t | t-1} &= a_i \mu_{t-1|t-1} + \lambda_i \mu_{z_i, t}^{(-i)} \\
    \sigma^2_{t|t-1} &= 
    a_i^2 \sigma^2_{t-1|t-1} + \sigma_x^2 + \lambda_i^2
    {\sigma^2}_{z_i,t}^{(-i)} \enspace.
    \label{supp-eq:predict_step_ep}
\end{align}  

\subsection{EP Moment Update}
After selecting a new cluster assignment $z_i$,
we update our likelihood approximation $\tilde{f}_i(\eta)$. 
We do this by selecting the parameters of $\tilde{f}_i(\eta)$ to minimize
the local KL divergence (Eq.~\eqref{eq:ep_update}) between
the \textit{tilted} distribution $\tilde{p}(\eta | z)$
\begin{equation*}
    \tilde{p}(\eta | z) \propto f_i(z_i, \eta) q(\eta|z_{-i}) = 
    \Pr(y_i \, | \, z_i, \eta) q(\eta | z_{-i}) \enspace
\end{equation*}
and the approximate distribution
$q(\eta|z) \propto \tilde{f}_i(\eta)q(\eta|z_{-i})$.
For Gaussian approximations (and more generally exponential families), 
minimizing the KL divergence is equivalent
to matching the expected values of $q(\eta | z)$'s sufficient statistics.
Because our approximation is a diagonal Gaussian,
its sufficient statistics are the marginal means and variances at each time
point $t$.

Therefore, we learn parameters of $q(\eta|z)$ to match the marginal means
and variances of $\tilde{p}(\eta_t | z)$ and 
then solve for $\tilde{f}_i$ by removing the cavity
distribution $q(\eta|z_{-i})$ from $q(\eta|z)$ in a similar manner to
Eqs.~\eqref{supp-eq:cavity_ts_mean} and \eqref{supp-eq:cavity_ts_var}.

Finally, the marginal mean and variance of 
the tilted distribution $\tilde{p}(\eta|z)$ can be efficiently calculated
using the forward and backward messages passed in
the Kalman smoother~\cite{bishop2006pattern}.

The forward message is
the filtered distribution of $x_t$ from the Kalman filter
\begin{equation*}
    \alpha(x_t) = \Pr(x_t \, | \, y_{1:t})
                \propto \Pr(y_{1:t}, x_t )
                =\int \Pr(y_t \, | \, x_t) \Pr(x_t \, | \, x_{t-1})
                \alpha(x_{t-1}) \ d x_{t-1}
\end{equation*}
where
\begin{equation*}
    \Pr(x_{t} \, | \, x_{t-1}) = \int \Pr(x_{t} \, | \, x_{t-1}, \eta_{t})
    q(\eta_t) \ d\eta_t \enspace.
\end{equation*}

The backward message is
the likelihood of future observations
\begin{equation*}
    \beta(x_t) = \Pr(y_{t+1:T} \, | \, x_t) 
    =\int \beta(x_{t+1}) \Pr(y_{t+1} \, | \, x_{t+1}) 
    \Pr(x_{t+1} \, | \, x_t) \ d x_{t+1} \enspace.
\end{equation*}

Then, the marginal distribution at time $\tau$ of $\eta_\tau$ is
\begin{align}
    \nonumber
    \tilde{p}(\eta_\tau \, | \, z) \propto& 
    \int \Pr(y_i \, | \, z_i, \eta) q(\eta | z_{-i}) \ d\eta_{-\tau} \\
    \nonumber
    =& \int \left[ \int \prod_t \Pr(y_{i,t} | x_{i,t}) 
        \Pr(x_{i,t} | x_{i,t-1}, \eta_{z_i, t}) \, dx_i  \right] 
        q(\eta|z_{-i}) \, d\eta_{-\tau}\\
    \nonumber
    =& \int \Big[
    \prod_{t < \tau} \Pr(y_{i,t} | x_{i,t}) 
        \Pr(x_{i,t} | x_{i,t-1}, \eta_{z_i, t}) 
        q(\eta_{z_i,t} | z_{-i}) \\
    \nonumber
    &\quad \times \ \bigg.
    \Pr(y_{i,\tau} | x_{i, \tau}) 
        \Pr(x_{i,\tau} | x_{i,\tau-1}, \eta_{z_i, \tau}) 
        q(\eta_{z_i,\tau} | z_{-i}) \\
    \nonumber
    &\quad \times \
    \prod_{t > \tau}  \Pr(y_{i,t} | x_{i,t}) 
        \Pr(x_{i,t} | x_{i,t-1}, \eta_{z_i, t}) 
        q(\eta_{z_i,t} | z_{-i}) 
    \Big] \ dx_i d\eta_{-\tau} \\
    \label{supp-eq:ts_moment}
    =& \int
    \alpha(x_{i,\tau-1}) \times
    \Pr(y_{i,\tau} | x_{i, \tau}) 
        \Pr(x_{i,\tau} | x_{i,\tau-1}, \eta_{z_i, \tau}) 
        q(\eta_{z_i,\tau} | z_{-i}) \times
    \beta(x_{i,\tau}) \ dx_{i,\tau-1} dx_{i,\tau} \enspace.
\end{align}
All terms within the integral on the final line of Eq.~\eqref{supp-eq:ts_moment}
are Gaussian.
Integrating out $x_{i,\tau}$ and $x_{i,\tau-1}$, gives us the tilted marginal
distribution for $\eta_{z_i,\tau}$.


Thus we can calculate the mariginal means and variances by passing the same 
messages as the univariate Kalman smoother in $O(T)$ time.


\section{EP Convergence}
\label{app:ep_converge}
This document outlines convergence analysis of our approximate collapsed
EP sampler.
We first review standard EP's convergence guarantees and
its dual representation (leading to a convergent inner-outer optimization).
We then bound the error for our approximations when performing our sampler.

Our task is to analyze the approximation accuracy of our EP approximation
$q^{(t)}$ for the posterior $p(\phi \ | \ y, z)$.

\subsection{Notation}
This subsection is for reference and can be skipped.

Variables:

\begin{itemize}
    \item $y = \{ y_i \}_{i = 1}^n$ are the observations
    \item $z = \{ z_i \}_{i = 1}^n$ are the latent cluster assignments
    \item $\phi = \{ \phi_k \}_{k=1}^K$ are the cluster parameter to collapse
\end{itemize}

Cluster Parameter Posteriors:

\begin{itemize}
    \item $p(\phi \ | \ y, z) = \prod_{k = 1}^K p(\phi_k \ | \ y, z)$
        is the conditional posterior of $\phi$ for assignment $z$
    \item $q^{(t)}(\phi) = \prod_{k = 1}^K q^{(t)}(\phi_k)$ is the
        approximation at time $t$
    \item $q^*(\phi \ | \ z) = \prod_{k = 1}^K q^*(\phi_k \ | \ z)$
        be the `optimal' exponential family approximation for assignment $z$
\end{itemize}

Likelihood/Site Approximations:

\begin{itemize}
    \item $\ell_i(\phi \ | \ z_i) = p(y_i \ | \ \phi, z_i)$ be the likelihood
        of observation $i$
    \item $\tilde{\ell}_i^{(t)}(\phi)$ be the likelihood (`site')
        approximation at time $t$
    \item $\tilde{\ell}^{*}_i(\phi \ | \ z)$ be the `optimal' likelihood
        approximation for assignment $z$
\end{itemize}

Exponential Family Parameters:

\begin{itemize}
    \item $\theta_k^{(t)}$ be the parameters of $q^{(t)}(\phi_k)$
    \item $\bar{\theta_k}(z)$ be the parameters of $q^*(\phi_k \ | \ z)$
    \item $\lambda_i^{(t)}$ be the parameters of the site approximation
        $\tilde{\ell}_i^{(t)}(\phi)$
    \item $\bar\lambda_i(z)$ be the parameters of the optimal site approximation
        $\tilde{\ell}_i^{*}(\phi \ | \ z)$
\end{itemize}

\subsection{Review of Standard EP's Theory}
The goal of Expectation Propagation (EP) is to find a distribution
$q$ restricted to exponential family $\mathcal{Q}$
\begin{equation}
    \mathcal{Q} = 
    \{ q(\phi) = \exp(\theta \cdot s(\phi) - A(\theta)) \ | \
        \theta \in \Theta\}
\end{equation}
such that it minimizes the KL-divergence from a target posterior $p(\phi | y)$
\begin{equation*}
    q^*(\phi) = \argmin_{q \in \mathcal{Q}} D_{KL}( p(\phi | y) \ || \ q(\phi))
\end{equation*}
This is accomplished by approximating likelihood terms with `site'
approximations
\begin{align}
    p(\phi | y) &\propto \pi(\phi) \prod_{i = 1}^n \ell_i(\phi) \\
    q(\phi) &\propto \pi(\phi) \prod_{i = 1}^n \tilde{\ell}_i(\phi)
    \enspace,
\end{align}
where $\tilde{\ell}_i \propto \exp(\lambda_i \cdot s(\phi))$ (without the
restriction that $\lambda_i \in \Theta$ or that $\tilde{\ell}_i$ can be
normalized).

The site approximations are calculated by projecting the `tilted' or hybrid
distribution and removing the `cavity' distribution
\begin{equation}
    \tilde{\ell}_i = \frac{\argmin_{q \in \mathcal{Q}} D(\tilde{p}_i ||
    q)}{q_{\backslash i}}
    \label{supp-eq:ep_update}
\end{equation}
where $q_{\backslash i}$ is the cavity distribution
\begin{equation*}
    q_{\backslash i}(\phi) \propto \pi(\phi) \prod_{j \neq i} \tilde{\ell}_j(\phi)
\end{equation*}
and $\tilde{p}_i$ is the tilted distribution
\begin{equation*}
    \tilde{p}_i \propto q_{\backslash i}(\phi) \ell_i(\phi) \propto
        \pi(\phi) \prod_{j \neq i} \tilde{\ell}_j(\phi) \cdot \ell_i(\phi)
    \enspace.
\end{equation*}

Standard EP works, by applying Eq.~\eqref{supp-eq:ep_update} until convergence.
However, standard EP is not guaranteed to converge and may have multiple fixed
points. To understand why this happens, its useful to consider the optimzation
problem EP implicitly solves.
By minimizing the KL-divergence to the tilted-distribtions, the fixed points of
EP are equivalent to maximizing the log-marginal probability using the Bethe
entropy approximation~\cite{teh2015distributed,heskes2002expectation}
\begin{equation}
    \argmax_{q \in \mathcal{Q}} -H(q) + \sum_{i = 1}^T (-H(\tilde{p}_i) + H(q))
    \enspace , 
\end{equation}
where $H(\cdot)$ is entropy.

This objective is not concave in $q$ (allows for multiple fixed points).
Furthermore, because EP applies the (simple to compute) "coordinate-ascent"
update (Eq.~\eqref{supp-eq:ep_update}), it's possible to fall into limit-cycles.

To overcome these problems, Heskes and Zoeter introduced an inner-outer
"double"-loop algorithm by optimizing the equivalent to the dual problem
\begin{equation}
    \max_{\theta} \min_{\{\lambda_i\}_{i=1}^n}  
    A\left(\lambda_0 + \sum_{i = 1}^n \lambda_i \right) + 
    \sum_{i = 1}^n \left( A_i(\theta - \lambda_i, 1) - A(\theta) \right)
\end{equation}
where $\theta$ are the parameters of the global approximation $q$,
$\lambda_i$ are the parameters of the site approximations $\tilde{\ell}_i$
and $A_i$ is the log-partition function of the tilted distribution
$\tilde{p}_i$.

This dual problem is concave in the site approximation parameters $\lambda$ and
by taking damped updates, it guaranteed to converge to a local optima.
The problem is that because this is a saddle-point problem (min $\lambda$, max
$\theta$), the (correct) outer loop updates to $\theta$ requires waiting until
$\lambda$ converge.
This was further extended to allow for distributed/parallel computation using
stochastic natural gradients by Teh et al.~\cite{teh2015distributed}.

Finally, EP has recently been shown to be consistent and exact in the large
data limit for the Gaussian approximating family~\cite{dehaene2015expectation}.
This was done by showing standard EP asymptotically behaves like the CCG
(Laplace approximation) of Newton-method's method iterates to the mode.

Solving EP's convergence and fixed point issues is a paper in itself;
however, we can show that the error between our Gibbs sampler EP-approximation
is not far off from what would happen if we ran EP to convergence after each
step of our Gibbs sampler.

\subsection{Sampling Gibbs}

We now consider our case where our target distribution is changing $p^{(t)}$,
as it depends on the sampled assignment $z^{(t)}$,
$p^{(t)} = p(\phi \ | \ y, z^{(t)}$).
We first review what happens at each iteration and then discuss error bounds.

Suppose $q^{(t)}(\phi) = \prod_{k = 1}^K \exp(\theta_k^{(t)} \cdot s(\phi_k) -
A(\theta_k^{(t)}))$ is our approximation for $p(\phi \ | \ y, z^{(t)})$.
Our sampling algorithm proceeds as follows:
\begin{enumerate}
    \item Select a latent assignment $z_i^{(t)}$ to reassign
    \item Calculate the cavity distribution $q^{(t)}(\phi \ | \ z_{-i}^{(t)})$,
        \begin{equation*}
            \theta_k^{\backslash i} =
            \begin{cases}
                \theta_k^{(t)} - \lambda_i^{(t)} & \text{ if } k = z_i^{(t)}\\
                \theta_k^{(t)} & \text{ otherwise}
            \end{cases}
        \end{equation*}
        where $\lambda_i^{(t)}$ are the parameters for site approximation $i$ 
    \item Approximate the collapsed likelihood for each $z_i$ assignment
        \begin{equation*}
            \Pr(y_i \ | \ z_i, z_{-i}^{(t)}, y_{-i}) =
            \int \Pr(y_i \ | \ z_i, \phi) q^{(t)}(\phi \ | \ z_{-i}^{(t)})
        \end{equation*}
    \item Sample a new $z_i^{(t+1)} = k$ proportional to the prior and collapsed likelihood
    \item Calculate the new site approximation $\lambda_i^{(t+1)}$
        \begin{align*}
            \lambda_i^{(t+1)} =
    \underbrace{
        \grad A^*(\grad_\theta A_i(\theta_k^{\backslash i}, 1))
    }_{\text{titled distribution projection}}
    - 
    \underbrace{(\theta_k^{\backslash i})}_{\text{cavity parameters}}
        \end{align*}
    \item Update the global approximations
        \begin{equation*}
            \theta_k^{(t+1)} =
            \begin{cases}
                \theta_k^{\backslash i} + \lambda_i^{(t+1)} & \text{ if }
                z_i^{(t+1)} = k \\
                \theta_k^{\backslash i} & \text{ otherwise}
            \end{cases}
        \end{equation*}
\end{enumerate}

Note that one step of our sampler only changes
$\theta_{z_i^{(t)}}$, $\theta_{z_i^{(t+1)}}$ and $\lambda_i$.

Outside of the iteration, we periodically (e.g. after one scan through the data)
run a full EP update without resampling a $z_i$.

\subsubsection{Error Bounds}
There are many types of error bounds that we could consider:
\begin{enumerate}[label=\textbf{(B\arabic*)}]
    \item $D(p_k (\phi_k | y, z^{(t)}) \ || \ q^{(t)}_k (\phi_k))$
        divergence between the exact posterior and our current approximation
    \item
        $D(q^*_k (\phi_k | y, z^{(t)}) \ || \ q^{(t)}_k (\phi_k))$
        divergence between the best and our current approximation
    \item
        $d(\bar{\theta}_k^{(t)},\, \theta_k^{(t)})$
        distance in terms of parameters
        $\theta_k \leftrightarrow q_k$
        between the best and our current approx
    \item 
        $D(\ell_i(\phi | z^{(t)}) \ || \ \tilde{\ell}_i^{(t)}(\phi))$
        divergence between the exact likelihood and our 
        current site approximations
    \item
        $D(\tilde{\ell}_i^{*}(\phi | z^{(t)}) \ ||\ \tilde{\ell}_i^{(t)}(\phi))$
        divergence
        between the best and our current site approximation
    \item
        $d(\lambda_i^*(z^{(t)}),\, \lambda_i^{(t)})$
        distance in terms of parameters 
        $\lambda_i \leftrightarrow \tilde{\ell}_i$
        between the best and our current site approx
\end{enumerate}

The first three quantities (B1-B3) (roughly) bound the global error between our current
approximation.
The last three quantities (B4-B6) (roughly) bound the local error of each site
approximation.
Note that the local bounds are stronger than the global bounds, as
the global parameter $\theta_k^{(t)}$ is the sum or local parameters
\begin{equation*}
    \underbrace{\theta_k^{(t)}}_{\text{global parameter}} = 
    \underbrace{\lambda_{0,k}}_{\text{prior term}} 
    + \underbrace{\sum_{i : z_i = k} \lambda_i}_{\text{local parameters}}
    \enspace.
\end{equation*}

\subsubsection{Global Approximation Bound}
Suppose $z_i$ was selected to be resampled at step $(t)$.
If $z_i$ does not change then, we have the standard EP update and its
convergence guarantees (or lack thereof).

If $z_i$ changes between time $(t)$ and $(t+1)$, then we can bound
the norm in term of parameters $d(\bar{\theta}_k,\, \theta_k)$ at time $(t+1)$
in terms of the norm at time $(t)$.
There are three cases depending on $k$:
\begin{enumerate}[label=\textbf{(\arabic*)}]
    \item If $k \neq z_i^{(t)}$ and $k \neq z_i^{(t+1)}$, then
        there were no changes to cluster $k$'s approximation or target
        (i.e. $\theta_k^{(t+1)} = \theta_k^{(t)}$ and
        $\bar\theta_k^{(t+1)} = \bar\theta_k^{(t)}$); therefore the error does
        not change
        \begin{equation}
            \underbrace{d(\bar{\theta}_k^{(t+1)},\,\theta_k^{(t+1)}) }_{
                \text{new error}} =
            \underbrace{d(\bar{\theta}_k^{(t)},\, \theta_k^{(t)}) }_{ 
            \text{old error}}
            \enspace.
        \end{equation}

    \item If $k = z_i^{(t)}$, then site $i$ is removed from cluster $k$
        (i.e. $\theta_k^{(t+1)} = \theta_k^{(t)} - \lambda_i^{(t)}$)
        and by applying the triangle equality we have
    \begin{align}
        \nonumber
        \underbrace{d(\bar{\theta}_k^{(t+1)},\,\theta_k^{(t+1)}) }_{
                \text{new error}}
        &=
        d(\bar{\theta}_k^{(t+1)},\, \theta_k^{(t)} - \lambda_i^{(t)})\\
        \nonumber
        &=
        d(\bar{\theta}_k^{(t+1)} + \lambda_i^{(t)},\, \theta_k^{(t)}) \\
        \nonumber 
        &\leq
        d(\bar{\theta}_k^{(t+1)} + \lambda_i^{(t)},\, \bar{\theta}_k^{(t)}) +
        d(\bar{\theta}_k^{(t)},\, \theta_k^{(t)}) \\
        &=
        d(\lambda_i^{(t)}, \, \bar{\theta}_k^{(t)} - \bar{\theta}_k^{(t+1)} ) +
            \underbrace{d(\bar{\theta}_k^{(t)},\, \theta_k^{(t)}) }_{ 
            \text{old error}}
       \label{supp-eq:global_bound_remove}
        \enspace,
    \end{align}
        therefore, the error increases by at most
        $d(\lambda_i^{(t)}, \, \bar{\theta}_k^{(t)} - \bar{\theta}_k^{(t+1)} )$
    how well $\lambda_i^{(t)}$ approximates the loss of $p(y_i | z_i, \phi)$
        in the optimal global approximation parameters.

    \item If $k = z_i^{(t+1)}$, then site $i$ is added to cluster $k$
        (i.e. $\theta_k^{(t+1)} = \theta_k^{(t)} + \lambda_i^{(t+1)}$)
        and by applying the triangle equality we have
    \begin{align}
        \nonumber
        \underbrace{d(\bar{\theta}_k^{(t+1)},\,\theta_k^{(t+1)}) }_{
                \text{new error}}
        &=
        d(\bar{\theta}_k^{(t+1)},\, \theta_k^{(t)} + \lambda_i^{(t+1)})\\
        \nonumber
        &=
        d(\bar{\theta}_k^{(t+1)} - \lambda_i^{(t+1)},\, \theta_k^{(t)}) \\
        \nonumber 
        &\leq
        d(\bar{\theta}_k^{(t+1)} - \lambda_i^{(t+1)},\, \bar{\theta}_k^{(t)}) +
        d(\bar{\theta}_k^{(t)},\, \theta_k^{(t)}) \\
        &=
        d(\lambda_i^{(t+1)}, \, \bar{\theta}_k^{(t+1)}- \bar{\theta}_k^{(t)} ) +
        \underbrace{d(\bar{\theta}_k^{(t)},\, \theta_k^{(t)}) }_{ 
        \text{old error}}
        \label{supp-eq:global_bound_insert}
        \enspace,
    \end{align}
        therefore, the error increases by at most
    $d(\lambda_i^{(t+1)}, \, \bar{\theta}_k^{(t+1)} - \bar{\theta}_k^{(t)})$.
    how well $\lambda_i^{(t+1)}$ approximates the addition of
    $p(y_i | z_i, \phi)$
        in the optimal global approximation parameters.
\end{enumerate}

In summary, the global approximation parameter error only grows in the two
cluster changed and the increase in error depends only on 
how well $\lambda_i^{(t)}$ approximates the loss of $p(y_i | z_i, \phi)$ 
in cluster $z_i^{(t)}$ (Eq.\eqref{supp-eq:global_bound_remove}) and 
how well $\lambda_i^{(t+1)}$ approximates the increase of $p(y_i | z_i, \phi)$ 
in cluster $z_i^{(t+1)}$ (Eq.\eqref{supp-eq:global_bound_insert}).

\subsection{Empirical Experiments}
The section describes a series of experiments to quantify the error induced by
only updating local sites compared against running full EP at each iteration.
For this experiment, we consider components that are GSM.
\begin{align}
    \nonumber
    p(y \ | \ \phi_k) &= (1-r)\cdot\mathcal{N}(y \ | \ \phi_k, \sigma^2) +
        r\cdot \mathcal{N}(y \ | \ \phi_k, C\sigma^2) \\
        &= f(y \ | \ \phi_k, r, C, \sigma^2) \enspace.
\end{align}

We measure the distance between our approximation $q^{(t)}$ and
$q^*$ at time $(t)$ (by running EP to convergence when $z^{(t)}$ is fixed)
using KL divergence and the percent error of recovering the posterior means and
variances for $\phi$.

In our experiments, 
we vary the \textit{proportion} probability $r$ from $[0, 0.5]$, 
the \textit{mean difference} $\Delta = (\phi_1-\phi_2)/\Var(y)$, 
and \textit{scale ratio} $C$.
Varying $r$ and $C$ determines how difficult the likelihood is to approximate
with a site approximation, while varying $\Delta$ determines how rapidly $z$
changes.
When $r = 0$, the problem is conjugate, so the error is zero and when $\Delta$
is large, $z$ rarely changes.

In all cases we find the error incurred by only using local updates does indeed level off 
(e.g. does not grow unbounded)
as there number of iterations increase. 
Furthermore this size of this error depends on the setting $r, \Delta, C$.

\begin{figure}[htb]
    \vskip 0.1in
    \centering
        \begin{minipage}[t]{.8\textwidth}
        \centering
            \includegraphics[width=\textwidth]{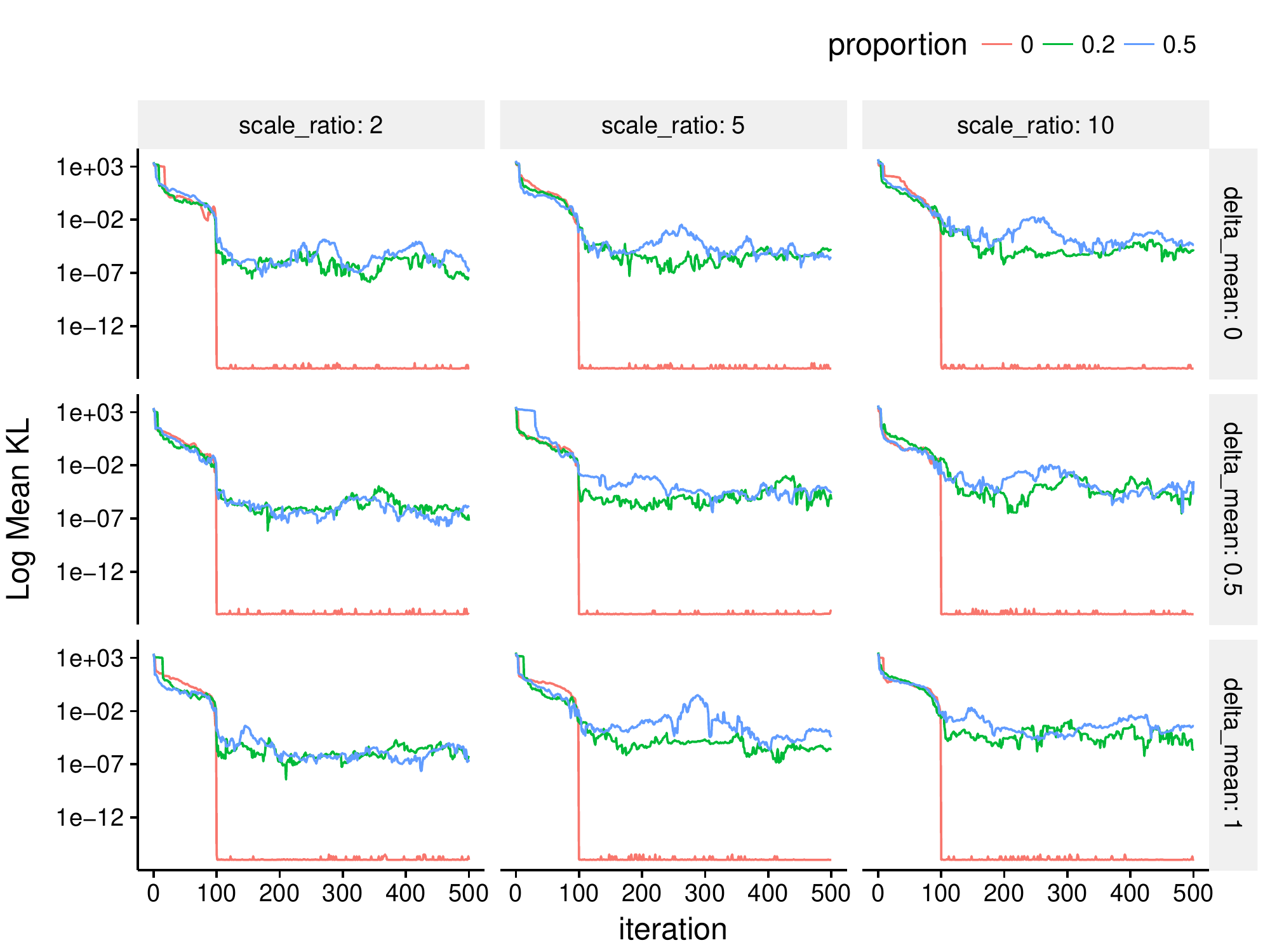}
        \end{minipage}
        
    \caption{KL error starting with flat site approximations.}
    \label{supp-fig:app_without_ep_conv}
    \vspace{0.1in} 
    \centering
        \begin{minipage}[t]{.8\textwidth}
        \centering
            \includegraphics[width=\textwidth]{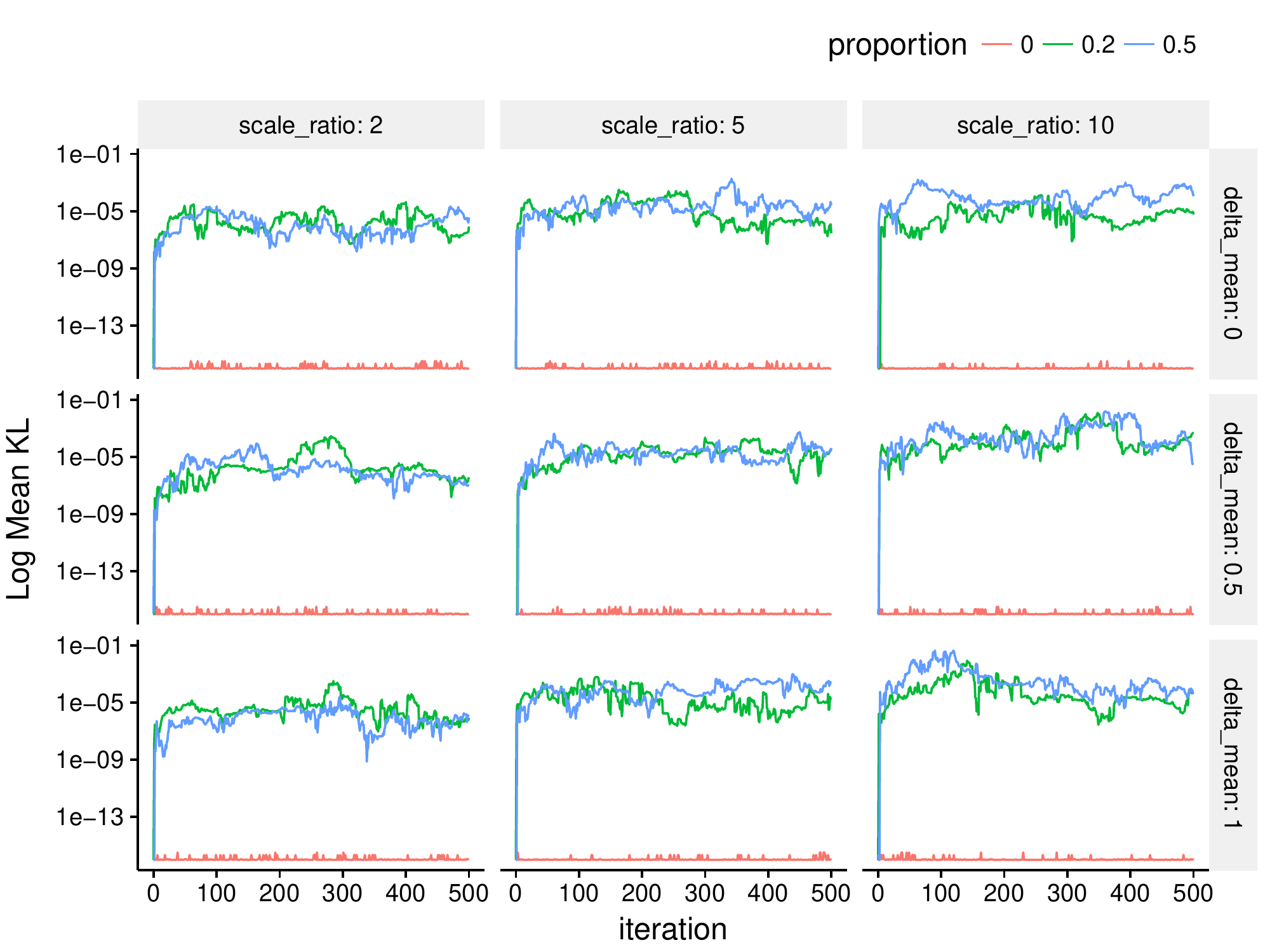}
        \end{minipage}
        
    \caption{KL error starting with site approximations from full EP.}
    \label{supp-fig:app_with_ep_conv}
    \vskip -0.1in

\end{figure}

Fig.~\ref{supp-fig:app_without_ep_conv} presents KL results for $n = 100$
when starting the site approximations from the prior (i.e. flat).
Note that after one pass, the approximation roughly level off (this includes
the setting of $\Delta = 0$, where $z$ is rapidly changing).

Fig.~\ref{supp-fig:app_with_ep_conv} presents KL results for $n = 100$
when starting the site approximations from full EP.
In this case, the error grows until it levels off at the same constant KL as
starting from flat approximations.

Finally Tables.~\ref{supp-tab:ep_conv_tab_without_ep} and
\ref{supp-tab:ep_conv_tab_with_ep}, show the percent error and
absolute percent error of the mean and variance for $\phi$ when $\Delta = 0.5$.

\begin{table*}
    \centering
    \caption{Median Percent Error and Median Absolute Percent Error (1 = 1\%)
    for $q^{(t)}$'s mean and variance after a full pass through the data
    starting with flat site approximations.
    Standard deviation estimates are presented in parenthesis.}
    \label{supp-tab:ep_conv_tab_without_ep}
    \vskip 0.1in
\begin{tabular}{rllllll}
  \hline
 & $C$ & $r$ & MeanPE & MeanAPE & VarPE & VarAPE \\ 
  \hline
1 & 2 & 0 & 0 (0) & 0 (0) & 0 (0) & 0 (0) \\ 
  2 & 2 & 0.2 & 0.05 (0.12) & 0.02 (0.1) & 0.03 (0.42) & 0.14 (0.3) \\ 
  3 & 2 & 0.5 & 0.02 (0.16) & 0.04 (1.43) & 0.05 (0.3) & 0.15 (0.2) \\ 
  4 & 5 & 0 & 0 (0) & 0 (0) & 0 (0) & 0 (0) \\ 
  5 & 5 & 0.2 & -0.01 (1.81) & 0.3 (1.47) & 0.31 (0.87) & 0.5 (0.55) \\ 
  6 & 5 & 0.5 & 3.6 (70.48) & 0.21 (62.94) & -0.31 (0.85) & 0.57 (0.56) \\ 
  7 & 10 & 0 & 0 (0) & 0 (0) & 0 (0) & 0 (0) \\ 
  8 & 10 & 0.2 & 1.56 (14.55) & 0.35 (13.08) & 0.72 (2.8) & 1.04 (2.06) \\ 
  9 & 10 & 0.5 & -0.2 (3.74) & 0.6 (3.11) & 1.27 (3.95) & 1.85 (2.82) \\ 
   \hline
\end{tabular}
\end{table*}

\begin{table*}
    \centering
    \caption{Median Percent Error and Median Absolute Percent Error (1 = 1\%)
    for $q^{(t)}$'s mean and variance after a full pass through the data.
    starting with full EP site approximations.
    Standard deviation estimates are presented in parenthesis.}
    \label{supp-tab:ep_conv_tab_with_ep}
    \vskip 0.1in
\begin{tabular}{rllllll}
  \hline
 & $C$ & $r$ & MeanPE & MeanAPE & VarPE & VarAPE \\ 
  \hline
1 & 2 & 0 & 0 (0) & 0 (0) & 0 (0) & 0 (0) \\ 
  2 & 2 & 0.2 & 0.02 (1.32) & 0.02 (0.1) & 0 (0.24) & 0.14 (0.3) \\ 
  3 & 2 & 0.5 & 0.01 (0.05) & 0.04 (1.43) & 0.01 (0.23) & 0.15 (0.2) \\ 
  4 & 5 & 0 & 0 (0) & 0 (0) & 0 (0) & 0 (0) \\ 
  5 & 5 & 0.2 & -0.15 (0.53) & 0.3 (1.47) & 0.09 (0.69) & 0.5 (0.55) \\ 
  6 & 5 & 0.5 & 0.02 (1.84) & 0.21 (62.94) & -0.11 (2.11) & 0.57 (0.56) \\ 
  7 & 10 & 0 & 0 (0) & 0 (0) & 0 (0) & 0 (0) \\ 
  8 & 10 & 0.2 & 0.08 (1.46) & 0.35 (13.08) & -0.21 (1.69) & 1.04 (2.06) \\ 
  9 & 10 & 0.5 & -0.2 (8.4) & 0.6 (3.11) & 0.28 (3.52) & 1.85 (2.82) \\ 
   \hline
\end{tabular}
\end{table*}

\clearpage
\section{Synthetic Time Series Trace Plots}
\label{app:traceplots}
In this section we provide additional plots showing the trace plots of the
model parameters $A, \lambda, \sigma_x^2, \sigma_y^2, x$ for the synthetic data
experiments in Sec.~\ref{sec:tscluster}.

\begin{figure}[htb!]
    \vskip 0.1in
    \centering
        \begin{minipage}[t]{.45\textwidth}
        \centering
            \includegraphics[width=\textwidth]{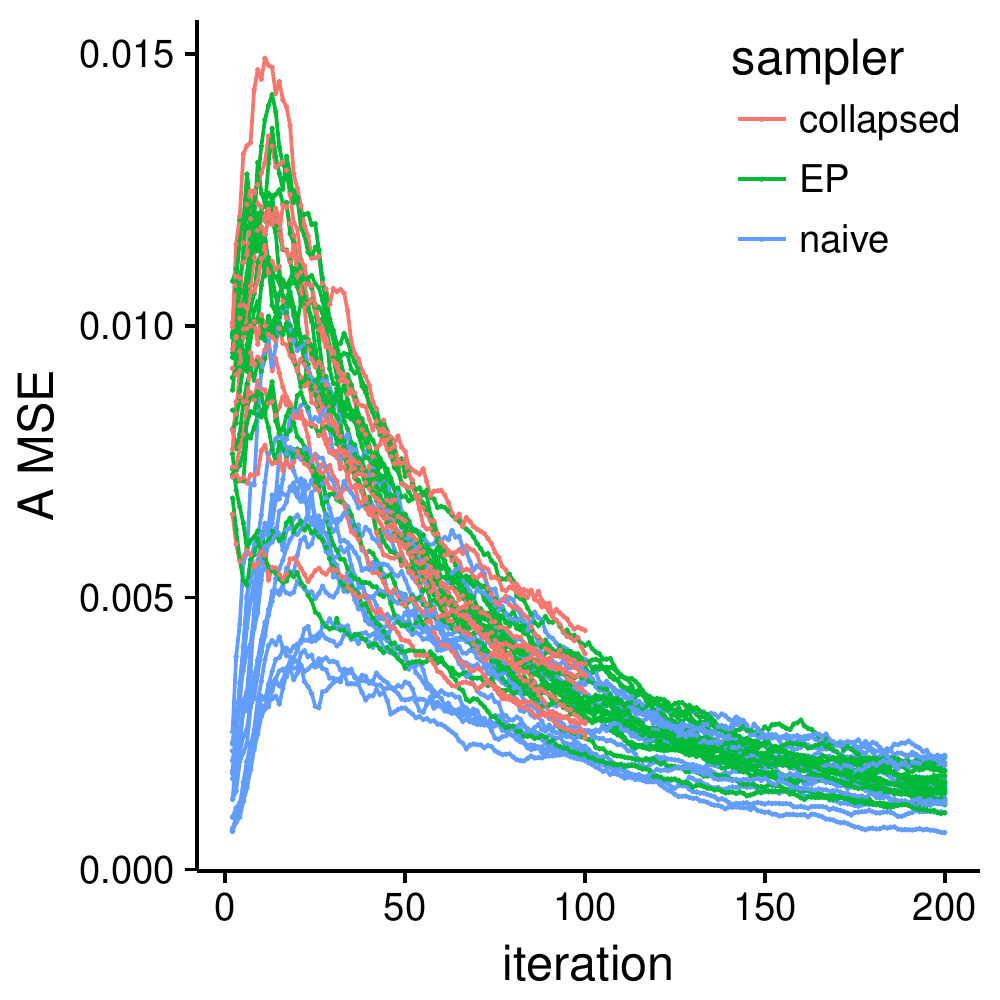}
        \caption*{(a) Trace of MSE $A$}
        \end{minipage}
        \hspace{0.1in}
        \begin{minipage}[t]{.45\textwidth}
        \centering
            \includegraphics[width=\textwidth]{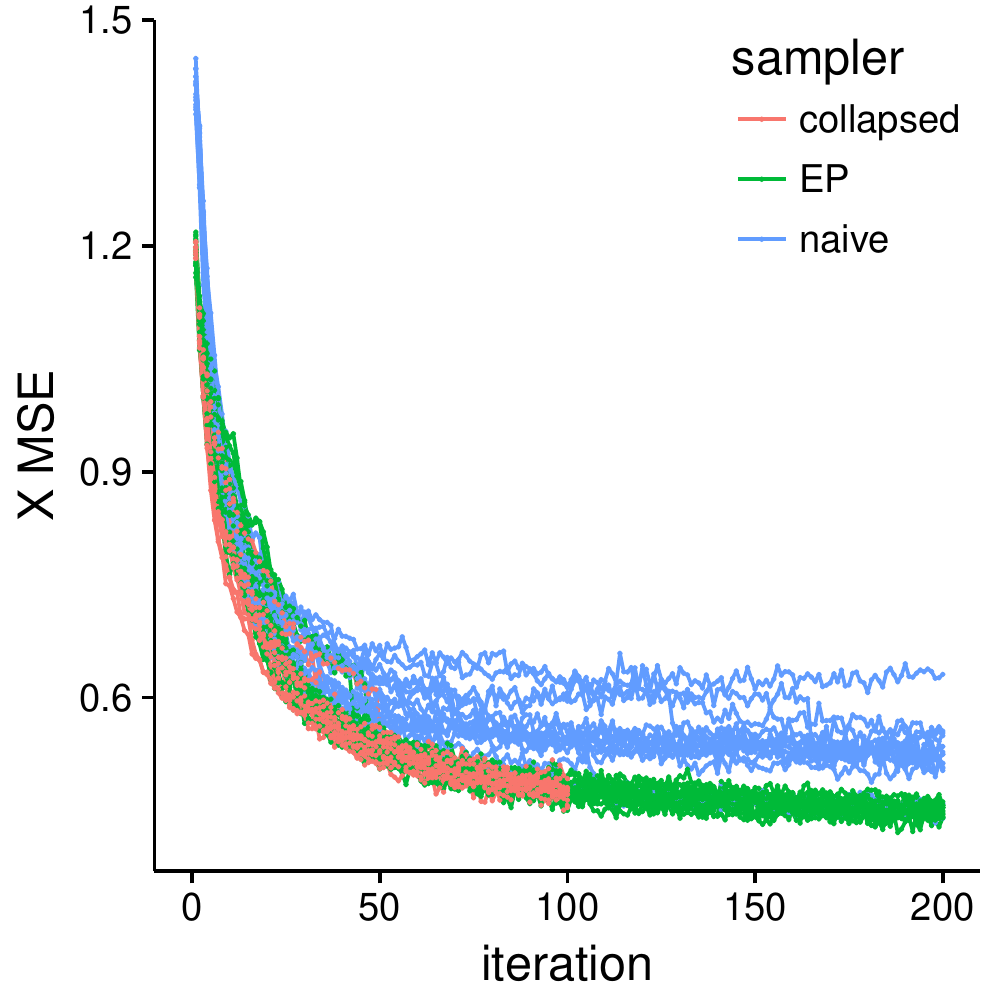}
            \caption*{(b) Trace of MSE $x$}
        \end{minipage}

        \begin{minipage}[b]{.45\textwidth}
        \centering
            \includegraphics[width=\textwidth]{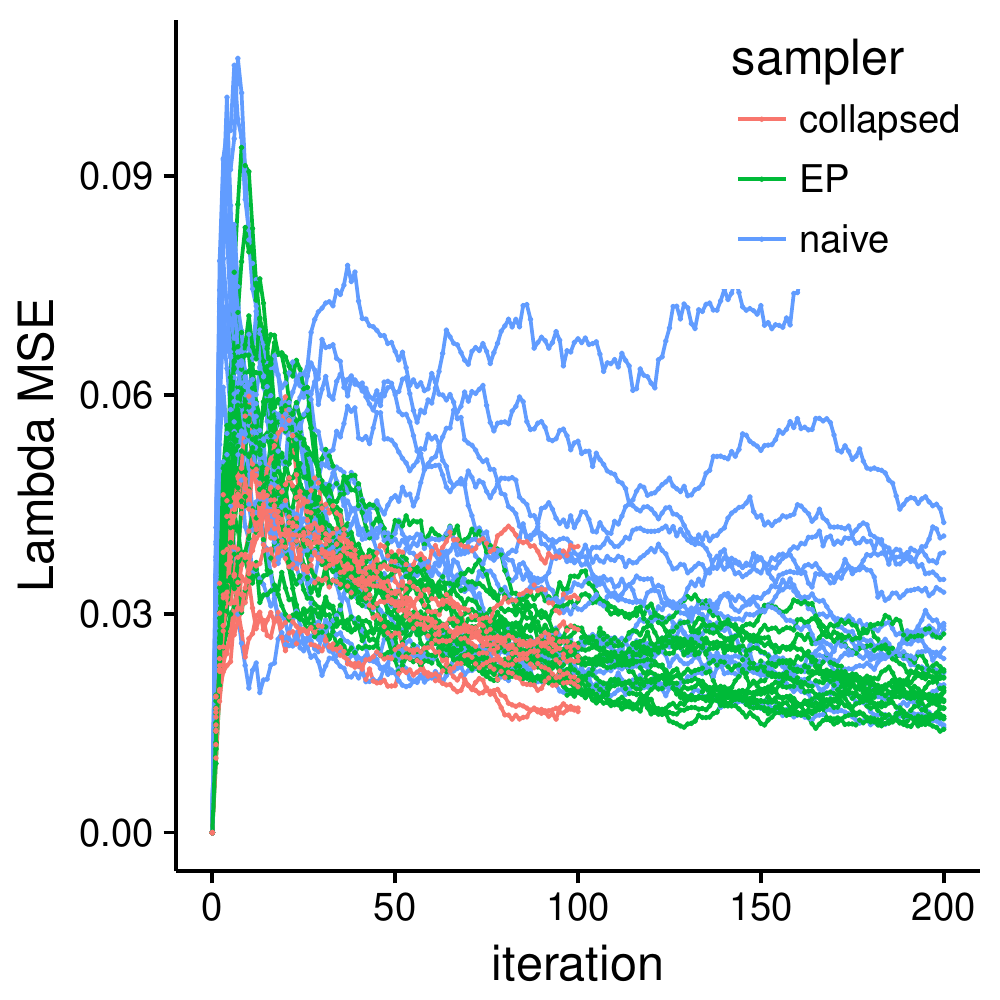}
        \caption*{(c) Trace of MSE $\lambda$}
        \end{minipage}
        \hspace{0.1in}
       \begin{minipage}[b]{.45\textwidth}
        \centering
            \includegraphics[width=\textwidth]{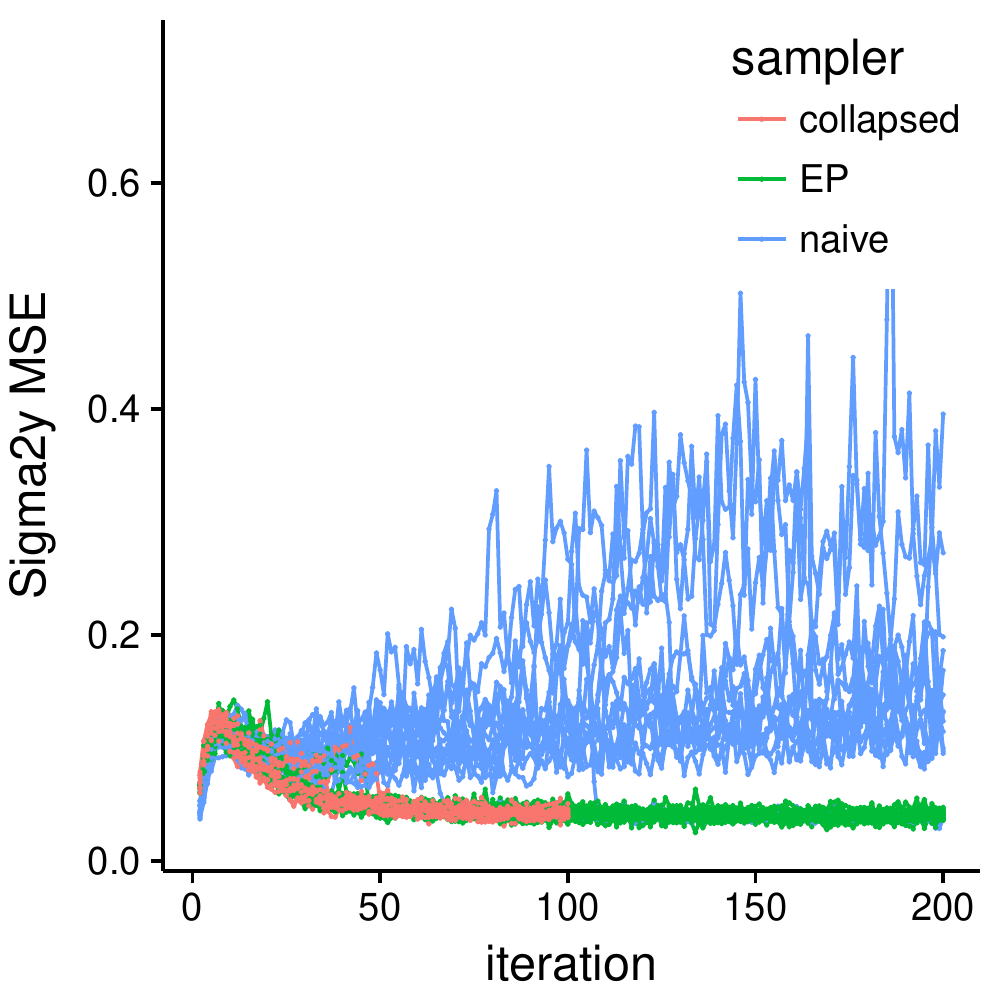}
        \caption*{(d) Trace of MSE $\sigma^2_y$}
        \end{minipage}
    \caption{MSE Traceplots of the synthetic timeseries samplers}
    \label{fig:app_traceplots}
    \vskip -0.1in

\end{figure}

\begin{figure}[h!]
    \vskip 0.1in
    \centering
        \begin{minipage}[t]{.65\textwidth}
        \centering
            \includegraphics[width=\textwidth]{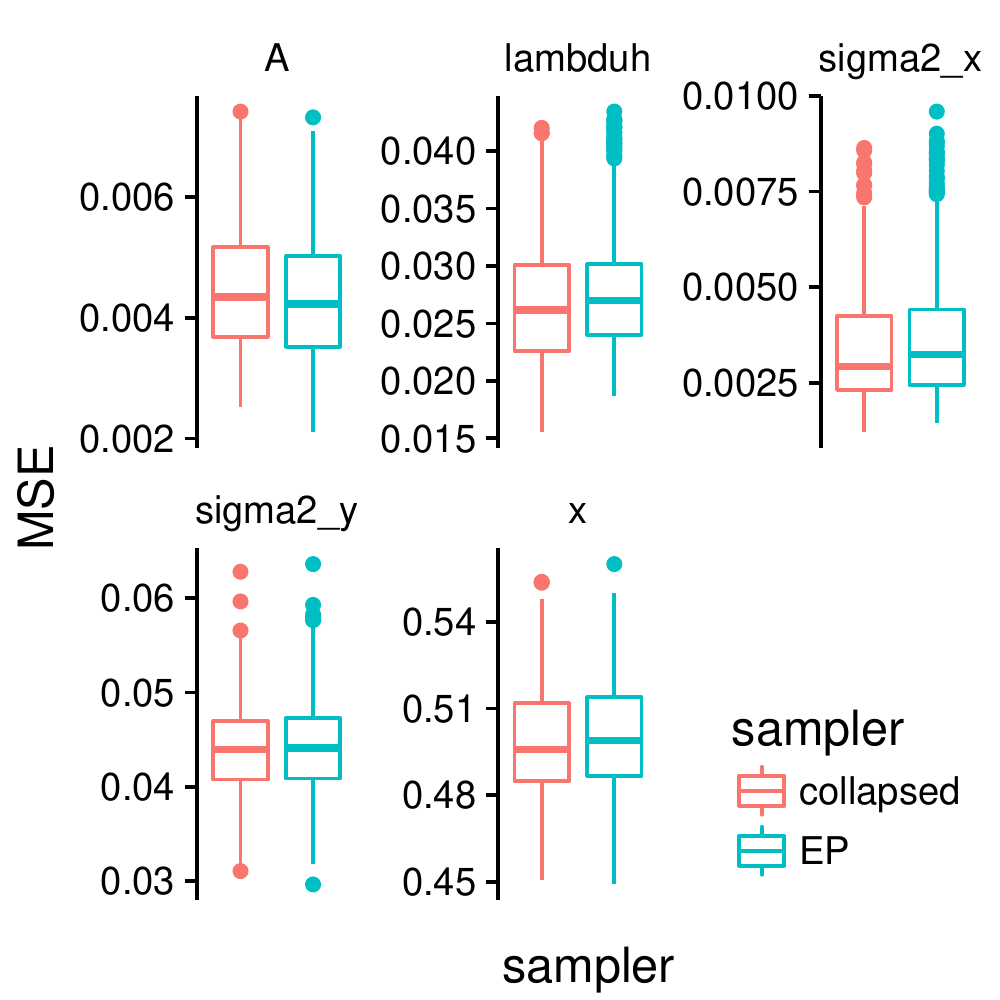}
        \end{minipage}
        
    \caption{Box-plot of posterior samples after burn-in}
    \label{fig:app_tsboxplot}
    \vskip -0.1in

\end{figure}

In Figs.~\ref{fig:app_traceplots}(a-d), 
we plot the mean squared error (MSE) between the sampled
parameter $\hat\theta$ and the true parameter $\theta^*$ of the synthetic data.
We can see that the collapsed sampler and our approximately collapsed EP
sampler have similar performance. 
In Fig.~\ref{fig:app_tsboxplot}, we plot box-plots comparing 'collapsed' and
'EP', showing it accurately estimates both the mean and variance.

\section{Seattle Housing Data}
\label{app:housing_data}

This section provides additional details for
the Seattle housing data example Sec.~\ref{sec:sea_housing_data}.

\subsection{Data Details}
We use the same dataset as Ren et al.~\cite{ren2015achieving}.
This consists of 124,480 transactions in 140 US census-tracts of the
city of Seattle from July 1997 to September 2013.
The time index for each transaction is at the monthly level,
therefore $T = 194$,
with multiple observations for in certain series-month pairs,
an no observations for other series-month pairs.

Each housing transaction contains the following house-specific covariates:
(i) number of bathrooms,
(ii) finished square-feet, and
(iii) lot-size square-feet.
We convert the house-specific covariates into feature variables by
taking their log-values and applying $B$-splines with knots at their quartiles.
Let $u_\ell$ denote the collection of features for house $\ell$.

\subsection{Housing Price Model}
To predict housing prices, we copy the model used by Ren et al.~\cite{ren2015achieving}
\begin{align}
    x_{i,t} &= a_i x_{i,t} + \lambda_i \eta_{z_i, t} + \epsilon_{i,t} \\
    \label{eq:housing_price_obs}
    y_{i,t,\ell} &= g_t + \beta_i u_\ell + x_{i,t} + \nu_{i,t,\ell}
    \enspace,
\end{align}
where $y_{i,t,\ell}$ denotes the log-price of house $\ell$ in region $i$ at
time $t$. 

The model for $y_{i,t, \ell}$, (Eq.~\eqref{eq:housing_price_obs}), consists
of four parts:
(i) a global housing price trend $g_{t}$ based on monthly seasonality,
(ii) a series-specific regression $\beta_i u_\ell$,
(iii) the latent residual process $x_{i,t}$,
(iv) white noise $\nu_{i,t,\ell}$.

The global trend $g_t$ is removed in a preprocessing step by the following
regression for parameters $\alpha_g$ and $\beta_g$
\begin{equation}
    \label{eq:global_reg}
    y_{i, t,\ell} \approx g_t = \alpha_g S(t) + \beta_g u_\ell \enspace,
\end{equation}
where $S(t)$ is a smooth spline basis over time $t$.
After learning $\alpha_g$ and $\beta_g$ in the preprocessing step,
the global trend $g_t$ is fixed.

After removing the global trend, the residual process is modeled as
the combination of region-specific regression and a latent AR(1) process.
Inference over $\beta_i$ and $x_{i,t}$ as well as all other model parameters
is achieved by Gibbs sampling.
Ren et al. provide the complete Gibbs sampling formulas~\cite{ren2015achieving}.

The difference between our two methods, collapsed and EP, is in how we sample
the series cluster assignments $z_i$.
For collapsed Gibbs, we run the expensive Kalman filter over individual clusters,
while for EP Gibbs, we use the approximate likelihood described in
Sec.~\ref{sec:tscluster} and Sec.~\ref{app:tscluster}.

\subsection{Additional Results}
We now present some diagnostics on the training data and the metrics of baseline models on the test data.

The other baseline models are:
\begin{itemize}
\item `global', the global trend $g_t$ from Eq.~\eqref{eq:global_reg}.
\item `{global+reg}', the global trend $g_t$ plus individual series-specific
        regression $\beta_i u_\ell$.
\end{itemize}

The metrics on the training data are presented in
Table~\ref{tab:train_housing}.
The metrics on the test data are presented in Table~\ref{tab:test_housing}.

In both cases, the algorithms using the time series clustering model (collapsed
and EP) vastly outperform the spline regression based models (`global' and
`global+regression').

\begin{table}
    \centering
    \caption{Training metrics on Seattle housing data for different algorithms
    averaged over 10 initializations.
    Parenthetical values are one standard deviation.}
    \label{tab:train_housing}
    \vskip 0.1in
    \begin{tabular}{lrrrrr}
      \hline
        metric & collapsed & EP & global & global+reg \\ 
      \hline
        RMSE & 119230 (150) & 119270 (220) & 205380 & 202050 \\ 
        Mean APE & 12.68 (0.01) & 12.69 (0.01) & 24.20 & 23.69 \\ 
        Median APE & 9.50 (0.01) & 9.49 (0.01) & 18.60 & 18.00 \\ 
        90th APE & 27.07 (0.01) & 27.1 (0.02) & 50.35 & 49.04 
    \\
      \hline
    \end{tabular}
\end{table}

\begin{table}
    \centering
    \caption{Test metrics on Seattle housing data for different algorithms
    averaged over 10 initializations.
    Parenthetical values are one standard deviation.}
    \label{tab:test_housing}
    \vskip 0.1in
    \begin{tabular}{lrrrrr}
      \hline
        metric & collapsed & EP & global & global+reg \\ 
      \hline
        RMSE & 125280 (50) & 125280 (80) & 182150 & 180285 \\ 
        Mean APE & 16.20 (0.01) & 16.20 (0.01) &24.20 & 23.55 \\ 
        Median APE & 12.07 (0.01) & 12.07 (0.01) & 18.59 & 18.17 \\ 
        90th APE & 34.17 (0.07) & 34.22 (0.05) & 50.48 & 49.31 \\
        Runtime & 121.6 (8.1) & 62.8 (3.7) & - & - \\
    \\
      \hline
    \end{tabular}
\end{table}



\end{document}